\documentclass[pmlr,twocolumn,10pt]{jmlr} 

\usepackage{caption}
\usepackage{subcaption}
\usepackage{graphicx}

\usepackage{booktabs}
\usepackage{siunitx}
\usepackage{physics}

\jmlrvolume{297}
\jmlryear{2025}
\jmlrworkshop{Machine Learning for Health (ML4H) 2025} 

\title[Synthetic Longitudinal EHRs for Chronic Disease]{Privacy-Preserving Generative Modeling and Clinical Validation of Longitudinal Health Records for Chronic Disease}

\author{\Name{Benjamin D. Ballyk\textsuperscript{1,2}}
       \Email{benjamin.ballyk@eng.ox.ac.uk} \\
       \Name{Ankit Gupta\textsuperscript{1}}
       \Email{ankit.m.gupta17@gmail.com} \\
       \Name{Sujay Konda\textsuperscript{1}}
       \Email{skonda1@terpmail.umd.edu} \\
       \Name{Kavitha Subramanian\textsuperscript{4}}
       \Email{kavithas@stanford.edu} \\
       \Name{Chris Landon\textsuperscript{5}}
       \Email{chris.landon@ventura.org} \\
       \Name{Ahmed Ammar Naseer\textsuperscript{1}}
       \Email{aanaseer22@gmail.com} \\
       \Name{Georg Maierhofer\textsuperscript{2,3}}
       \Email{gam37@cam.ac.uk} \\
       \Name{Sumanth Swaminathan\textsuperscript{1,2}}
       \Email{sswami@vironix.ai} \\
       \Name{Vasudevan Venkateshwaran\textsuperscript{1}}
       \Email{vvenkate@vironix.ai} \\ \\
       \addr \textsuperscript{1} Vironix Health Inc, Austin, TX, USA \\
       \addr \textsuperscript{2} University of Oxford, Oxford, UK \\
       \addr \textsuperscript{3} University of Cambridge, Cambridge, UK \\
       \addr \textsuperscript{4} Stanford University, Stanford, CA, USA \\
       \addr \textsuperscript{5} University of Southern California, Los Angeles, CA, USA \\
} 

\begin{document}

\maketitle

\vspace*{-4em}

\begin{abstract}

Data privacy is a critical challenge in modern medical workflows as the adoption of electronic patient records has grown rapidly. Stringent data protection regulations limit access to clinical records for training and integrating machine learning models that have shown promise in improving diagnostic accuracy and personalized care outcomes. Synthetic data offers a promising alternative; however, current generative models either struggle with time-series data or lack formal privacy guaranties. In this paper, we enhance a state-of-the-art time-series generative model to better handle longitudinal clinical data while incorporating quantifiable privacy safeguards. Using real data from chronic kidney disease and ICU patients, we evaluate our method through statistical tests, a Train-on-Synthetic-Test-on-Real (TSTR) setup, and expert clinical review. Our non-private model (Augmented TimeGAN) outperforms transformer- and flow-based models on statistical metrics in several datasets, while our private model (DP-TimeGAN) maintains a mean authenticity of 0.778 on the CKD dataset, outperforming existing state-of-the-art models on the privacy-utility frontier. Both models achieve performance comparable to real data in clinician evaluations, providing robust input data necessary for developing models for complex chronic conditions without compromising data privacy.

\end{abstract}

\paragraph*{Data and Code Availability}
The datasets used in the study are publicly available. The time-series eICU dataset originates from the eICU Collaborative Research Database created by the Philips eICU Research Institute, and is available after required PhysioNet credentialing \citep{pollard_eicu_2017, pollard_eicu_2018}. The chronic kidney disease dataset is pulled from the CKD-ROUTE study, which monitored patient prognosis over a three-year window \citep{iimori_data_2018, iimori_prognosis_2018}. The code is available at  \href{https://github.com/Vironix-Science/ppehcrgen}{https://github.com/Vironix-Science/ppehcrgen}.

\paragraph*{Institutional Review Board (IRB)}
No IRB approval was necessary for this project, as the data used is de-identified and publicly available.

\begin{figure*}[thb]
    \centering
    \includegraphics[width=1\linewidth]{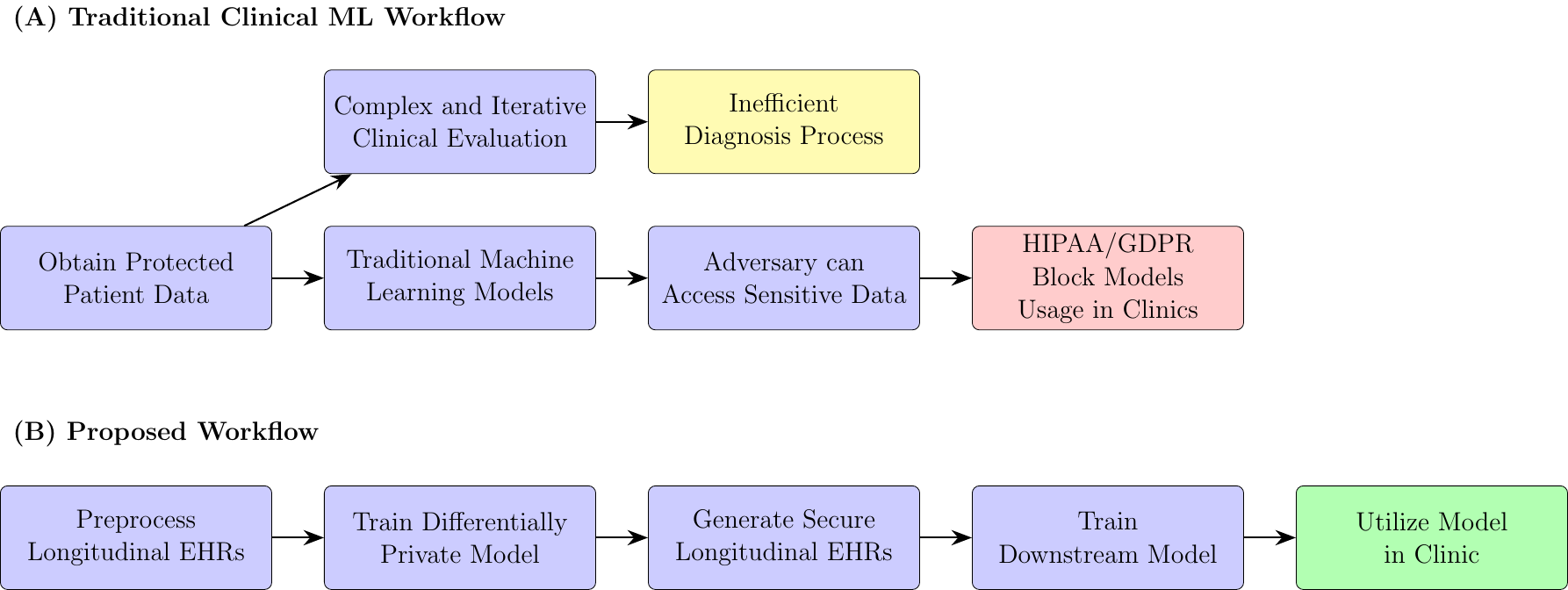}
    \caption{(A) Current workflow when handling protected patient data within the clinic. (B) Proposed downstream model pipeline for generic secure patient evaluation with machine learning models.}
    \label{fig:pipeline}
\end{figure*}

\section{Introduction}
\label{sec:intro}

Recent advancements in the utility and breadth of machine learning (ML) have unlocked several applications for enhancing and streamlining medical workflows. In particular, risk prediction, triage, disease progression modeling, and early detection are among many clinical tasks that have proven to be conducive to ML techniques \citep{law_machine_2019,mienye_improved_2020,swaminathan_machine_2017}. However, the integration of large-scale predictive and diagnostic models in clinical settings has been constrained by stringent privacy regulations.

Broad data privacy laws such as the Health Insurance Portability Accountability Act (HIPAA) in the United States and the General Data Protection Regulation (GDPR) in Europe are designed to protect medical patients from fraud and promote the adoption of electronic health records (EHRs) across medical institutions. Both HIPAA and GDPR mandate that hospitals minimize the quantity of data released, and often require explicit consent from patients prior to data disclosure \citep{accessed_4_6_25_european_parliament_and_council_of_the_european_union_general_2025,accessed_4_6_25_office_for_civil_rights_hipaa_2024}. Consequently, procuring EHRs has become costly and time-consuming for researchers and private stakeholders. Synthetic health records have recently gained traction as an avenue to address these challenges. 

The notion of synthetic data first emerged in the early 1990s as a statistical method to enable meaningful analysis without compromising individual privacy \citep{rubin_statistical_1993}. Unfortunately, early efforts to extend these concepts into the clinical domain were hindered by insufficient computing power, inconsistent data standards, and the low utility of synthetic outputs {\citep{gonzales_synthetic_2023}.

Over the past decade, advances in deep generative models and the widespread adoption of electronic health records (EHRs) have revitalized the practice of synthetic data generation \citep{chen_synthetic_2021, van_breugel_synthetic_2024, ktena_generative_2024}. Typically, diagnostic pipelines rely on either a complicated physical diagnosis process or ML models that use protected patient data for training. Recently, with the surge of synthetic data, the flow of clinical data begins by passing records to a generative model to be used either for training or direct modification. Then, the output may be used to train downstream ML models for medical prediction or classification tasks. (cf. Figure~\ref{fig:pipeline}).

While deep generative models have empirically excelled at generating static snapshots of clinical information \citep{foraker_spot_2021}, they have struggled to accomodate time-dependent (longitudinal) records necessary for the development of forward-predictive disease progression models. Moreover, the generative models available for time-series applications fail to produce quantifiable privacy controls to protect patient information.

In this paper, we introduce the Differentially Private TimeGAN (DP-TimeGAN) model, which incorporates differential privacy into the training processes of a generative adversarial network for quantifiable patient data security. 

\section{Related work and contributions}
\label{related_work}

Generating realistic longitudinal EHRs is challenging due to their high dimensionality, long sequence lengths, and frequent discontinuities. To address these challenges, recent work on generative models for time-series has explored several architectures with potential for synthesizing longitudinal EHRs. Temporal variational autoencoders (TimeVAEs) were designed to stabilize training, but struggle to capture the abrupt changes common in longitudinal EHRs \citep{desai_timevae_2021,kingma_auto-encoding_2022}. Conversely, temporal fusion transformers (TFTs) can represent long, irregular sequences \citep{lim2021temporal}, but their architectural backbone infamously suffers from time scaling quadratically with sequence length, which is prohibitive for quickly generating long longitudinal records \citep{sommers_survey_2024,vaswani_attention_2017}. Diffusion models have been adapted to quickly produce time-series data; however, they currently produce lower-fidelity results relative to other methods \citep{lin_diffusion_2023,sohl-dickstein_deep_2015}.

Generative adversarial networks (GANs) have shown to balance fidelity and computational efficiency, requiring relatively few parameters and generating results quickly in a single forward pass \citep{goodfellow_generative_2014}. Our investigation begins with TimeGAN, a widely used baseline for time-series synthesis \citep{yoon_time-series_2019}, which, in its default state, does not enforce privacy guarantees. We benchmark against the recent SeriesGAN \citep{eskandari_nasab_seriesgan_2024}, TransFusion \citep{sikder_transfusion_2025}, and TimeDiff \citep{tian_reliable_2024} models. 

\begin{figure}[h]
    \centering
    \includegraphics[width=0.9\linewidth]{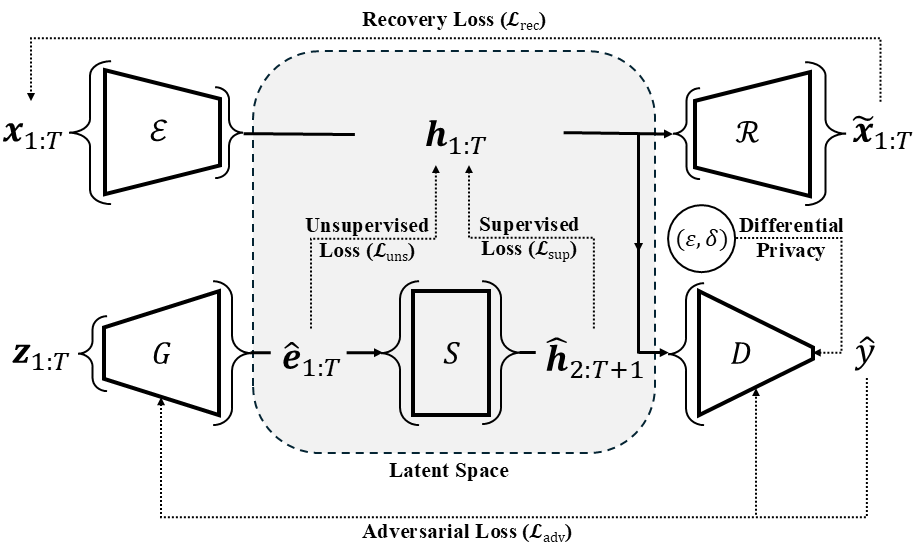}
    \caption{
    Model architecture for DP-TimeGAN. The model consists of five recurrent networks: embedding ($\mathcal{E}$), recovery ($\mathcal{R}$), supervisor ($\mathcal{S}$), generator ($G$), and discriminator ($D$). Real sequences $\mathbf{x}_{1:T}$ are mapped to latent space as $\mathbf{h}_{1:T} = \mathcal{E}(\mathbf{x}_{1:T})$. The generator produces latent sequences $\mathbf{\hat{e}}_{1:T} = G(\mathbf{z}_{1:T})$ from random noise, which are refined by the supervisor into supervised embeddings $\mathbf{\hat{h}}_{2:T+1} = \mathcal{S}(\mathbf{\hat{e}}_{1:T})$. The recovery network maps latent sequences back to data space, yielding $\mathbf{\tilde{x}}_{1:T} = \mathcal{R}(\mathbf{h}_{1:T})$, and the discriminator outputs $\hat{y}\in [0,1]$ as the classification of latent sequences for adversarial training.\vspace{-0.5cm}
    }
    \label{fig:model_architecture}
\end{figure}

Recent literature has also tested several strategies to enforce privacy requirements within deep generative models. Heuristic approaches, such as the identifiability loss in ADS-GAN, embed privacy into training, but lack reproducible, provable bounds \citep{yoon_anonymization_2020}. We instead emphasize differential privacy, which introduces noise and gradient clipping during training to track and bound cumulative privacy loss \citep{abadi_deep_2016,vaudenay_our_2006}. As a differentially private baseline, we also benchmark against Differentially Private Normalizing Flows (DP Normalizing Flows) \citep{lee_differentially_2022}.

\section{Methodology}
\label{methods}

Time series health records are characterized by chronologically ordered observations $\mathbf{x}_{1:T}=(\mathbf{x}_1, \mathbf{x}_2,\ldots, \mathbf{x}_T)$, where each data vector may exhibit a dependence on previous observations. Given $\{\mathbf{x}_{1:T}\}_{i=1}^{N} \sim \mathcal{D}$ for some unknown data distribution, an effective generative model must approximate $\mathcal{D}$ for sampling without purely replicating data samples. We accomplish this by augmenting a powerful recurrent generative adversarial network, and incorporating differentially private training (cf. Figure~\ref{fig:model_architecture}).

\subsection{Time-series Generative Adversarial Networks} \label{methods_timegan}
Our method for longitudinal EHR generation starts with the Time-series Generative Adversarial Network (TimeGAN), which comprises five recurrent neural networks (RNNs), working together to learn temporal dynamics in a latent space: the embedding ($\mathcal{E}$), recovery ($\mathcal{R}$), supervisor ($\mathcal{S}$), generator ($G$) and discriminator ($D$) networks. Real data sequences are denoted $\mathbf{x}_{1:T}$, and random noise sequences by $\mathbf{z}_{1:T}$. Synthetic data is obtained from $\mathbf{z}_{1:T}$ by passing the supervised latent sequences through the recovery network: $\tilde{\mathbf{x}}_{1:T} = \mathcal{R}(\mathcal{S}(G(\mathbf{z}_{1:T})))$.

TimeGAN is trained in three steps. Firstly, $\mathcal{E}$ and $\mathcal{R}$ are trained to compress raw data into a lower-dimensional latent space and reconstruct them back to feature space. This is achieved by minimizing the reconstruction loss,  $\mathcal{L}_{\text{rec}} = \mathbb{E}\!\left[\|\mathbf{x}_{1:T} - \mathcal{R}(\mathcal{E}(\mathbf{x}_{1:T}))\|^2\right]$.
 
Next, $S$ is trained to learn temporal dynamics in the latent space by performing next-step prediction:
$\mathcal{L}_{\text{sup}} 
= \mathbb{E}\!\left[\left\|\mathcal{E}(\mathbf{x}_{2:T+1}) - S(\mathcal{E}(\mathbf{x}_{1:T}))\right\|^2\right]$. This means $S$ enforces temporal consistency in the latent space. Notably, supervisory loss information continues to backpropagate through $\mathcal{E}$.

Finally, the generator ($G$) and discriminator ($D$) networks are trained adversarially using the typical min-max GAN objective, 
\begin{align*}
\begin{split}\mathcal{L}_{\text{adv}}=\min_{G} \max_{D} \;
&\mathbb{E}_{x \sim p_{\text{data}}(x)} \left[ \log D(\mathbf{x}_{1:T}) \right] \\
&+\mathbb{E}_{z \sim p_z(z)} \left[ \log \left( 1 - D(S(G(\mathbf{z}_{1:T}))) \right) \right]
\end{split}
\end{align*}

while $\mathcal{E}$ and $S$ continue to train. The adversarial loss is also computed on unsupervised embeddings, creating the unsupervised loss ($\mathcal{L_{\text{uns}}}$). Gated recurrent units (GRUs) are used as the default RNN architecture for all internal networks \citep[cf.][]{cho_learning_2014}. A schematic overview of TimeGAN is shown in Figure~\ref{fig:model_architecture}.

Below, we describe two further modifications to TimeGAN's training protocol which have been experimented with to improve the generation of synthetic EHRs while stabilizing training; we refer to the version with the highest performance as the `Augmented TimeGAN', and is used for further comparisons.

\subsubsection{Discriminator Noise Injection} \label{methods_noise_injection}

In practice, adversarial training in TimeGAN is often unstable as $D$ quickly outperforms $G$ in early training. This imbalance is particularly pronounced in EHRs due to the complex temporal dependencies and sparse observations. 

To optimize generator expressivity, we inject Gaussian noise into discriminator ground-truth inputs, 

\begin{equation}
\begin{aligned}
    \hat{y}_{\text{real}} &= D\!\left( \mathbf{h}_{1:T} + \mathbf{n}_{1:T} \right), \\
    \mathbf{n}_{t} &\sim \mathcal{N}(\mathbf{0}, \sigma^2 \mathbf{I}), 
    \quad t = 1,\ldots,T.
\end{aligned}
\label{eqn: add noise change}
\end{equation}

Here, $\hat{y}_{\text{real}}$ are ground-truth discriminator outputs, $\mathbf{h}$ are real embeddings, $\mathbf{n}$ is the injected noise vector sequence, and $\sigma$ is the standard deviation of noise. This process regularizes $D$, slows early dominance, and ensures that stronger gradient signals reach $G$, leading to more realistic synthetic EHR sequences.

Figure \appendixref{fig:aug_sine_synthesis} shows an example of the synthetic results from training augmented TimeGAN on the sinusoidal dataset. We observe in Figures \appendixref{fig:realsine_paths} and \appendixref{fig:fakesine_paths} that synthetic temporal paths capture the dynamics of real data, and that their distributions are roughly aligned, as evidenced by the PCA and t-SNE visualizations shown in Figures \appendixref{fig:pca_sines} and \appendixref{fig:tsne_sines}.

\subsubsection{Extended Long Short-Term Memory Blocks} \label{methods_xlstm}

Longitudinal EHRs often contain extended sequences of observations. While GRUs in TimeGAN are effective for processing and replicating shorter sequences, their performance deteriorates on longer sequences \citep{bai2018empirical}. To address this limitation, we experiment with Extended Long Short-Term Memory (xLSTM) blocks in $G$, which capture long-range dependencies efficiently without costly autoregression \citep{beck_xlstm_2024}. However, in our ablation study (Table \ref{tab:ablation}), the 1:1 mLSTM : sLSTM blocks configuration did not improve generation quality, suggesting that standard GRUs suffice for the sequence lengths in our datasets. Therefore, the xLSTM block was not used in $G$ for the Augmented TimeGAN.

\subsection{Incorporating Differential Privacy} \label{methods_dp}

Differential privacy (DP) quantifies the extent to which a transformation mechanism releases information about individual records in a dataset. A mechanism $\mathcal{M}$ is said to be $(\varepsilon, \delta)$-differentially private if for any two adjacent datasets, $d$ and $d'$, 
\begin{align}
    P[\mathcal{M}(d) \in S] \leq e^{\varepsilon} P[\mathcal{M}(d') \in S] + \delta,
    \label{eqn: epsilon delta dp}
\end{align}

where, $S \subseteq \text{Range}(\mathcal{M})$. Here, the parameter $\varepsilon \geq 0$ limits the maximum change in the output probabilities; smaller values provide stronger privacy. The parameter $\delta$ is the tail risk, which represents the probability that the $\varepsilon$ privacy guarantee may be violated completely \citep{vaudenay_our_2006}. 

In EHR synthesis, differential privacy (DP) provides a quantifiable measure of privacy loss from real patient data. In all experiments, we set $\epsilon \in [10, 20]$ and $\delta = 10^{-5}$, which is consistent with other large-scale government and personal data releases (Table~\appendixref{tab:epsilon_values}). 

DP in machine learning relies on three core mechanisms to limit privacy leakage \citep{abadi_deep_2016}: (i) gradient clipping, which bounds the contribution of any individual sample; (ii) noise injection, which obscures aggregated gradients to further reduce single-sample influence and facilitates privacy accounting; and (iii) random batch sampling, which selects samples independently in each batch, preventing correlations between patient records from being learned.

As training proceeds, privacy loss accumulates with each optimization step. This is tracked via a privacy accounting framework, summing per-epoch $\epsilon$ to ensure it remains below the predefined budget. In our experiments, DP was implemented in the discriminator ($D$) using Opacus \citep{xie_differentially_2018,yousefpour_opacus_2022} with Renyi differential privacy employed for accounting \citep{mironov_renyi_2017}. This incorporation of DP into the Augmented TimeGAN thus creates the DP-TimeGAN, selected for consistency of outputs.

\section{Evaluation}
\label{eval}
Evaluating synthetic longitudinal data is challenging, as multivariate sequences do not readily lend themselves to traditional cross-sectional statistical analyses \citep{alaa_how_2022,dankar_multi-dimensional_2022}. To address this, we assess data quality using four key characteristics, complemented by end-use evaluations to link quantitative results to clinical relevance.

\subsection{Fidelity, Diversity, and Privacy Metrics}
\label{eval_fidelity_diversity_privacy}

\textbf{Fidelity} measures the plausibility of synthetic data relative to real patient EHRs. We measure fidelity using three metrics in our experiments. First, \textbf{Maximum mean discrepancy (MMD)} measures the distributional distance between real and synthetic data using Gaussian kernels \citep{xu_empirical_2018}. An additional measure of fidelity is \textbf{$\alpha$-precision}, which is based on minimum volume sets, and evaluates the overlap between the majority of real data and the synthetic distribution, discounting outliers \citep{alaa_how_2022}. Finally, we also measure fidelity via a \textbf{discriminative score (DS)}, which measures the accuracy of a post-hoc GRU-based discriminator network in distinguishing real from synthetic EHRs \citep{yoon_time-series_2019}.

\textbf{Diversity} assesses whether synthetic data capture the full variability of the real dataset. We measure diversity with two strategies: First, the \textbf{$\beta$-recall}, the overlap between the minimum volume set of synthetic data and the full real data distribution \citep{alaa_how_2022}. Secondly, we use \textbf{Principal Component Analysis (PCA)} and \textbf{t-Stochastic Neighbor Embedding (t-SNE)} plots, which are dimensionality reduction techniques that allow visual assessment of distributional alignment of high-dimensional data using low-dimensional projections \citep{goar_data_2021,JMLR:v9:vandermaaten08a}.

\textbf{Privacy} metrics are necessary to validate that patient data remains quantifiably secure. The \textbf{authenticity metric} evaluates the fraction of synthetic samples that are not close to any real training sample, thereby indicating the model's generalization capability and mitigating the risk of overfitting \citep{alaa_how_2022}.

Further details on our metrics and their mathematical formulations are included in \appendixref{appendix_metric}.

\subsection{Downstream Utility} \label{eval_downstream_utility}

Downstream utility quantifies the practical usefulness of synthetic data in real predictive tasks, evaluated using a ``Train on Synthetic, Test on Real'' (TSTR) framework. We measure utility with the \textbf{predictive score (PS)}, defined as the mean absolute error of a post-hoc GRU predictor trained on synthetic data to forecast the next time step of real EHR sequences \citep{yoon_time-series_2019}.

Additionally, to assess applicability for chronic disease modeling, we also implement a \textbf{downstream classification task} using synthetic EHRs from the CKD dataset. A GRU-based classifier is trained on diabetes flags in synthetic data and evaluated on real patient data under the TSTR setup, with performance quantified by the \textbf{Area Under the Receiver Operating Characteristic curve (AUC-ROC)}. Further details are provided in \appendixref{appendix_utility_metrics}.

\subsection{Blinded Clinician Validation}
\label{eval_clinician_evaluation}
Beyond statistical similarity, synthetic longitudinal EHRs must exhibit clinically credible trajectories. For blinded validation, we randomly select 25 CKD patient profiles, comprising a mix of real data and synthetic outputs of each generative model. Each profile includes: (i) patient age and gender, (ii) baseline measurements of body mass index (BMI), hemoglobin (Hb), albumin (Alb), creatinine (Cr), and urinary protein-to-creatinine ratio (UPCR), and (iii) a three-year sequence of estimated glomerular filtration rate (eGFR) recorded every 6 months. Example profiles are shown in (Appendix~\ref{appendix_clinician}). Profiles were evaluated by five CKD specialists, who answered \hyperref[list:clinician_evaluation_questions]{three evaluation questions} for each profile.

From responses to question 1, we calculate two realism metrics: \textbf{Relaxed R/U}, where a sample was deemed realistic if at least one clinician labeled it realistic, and \textbf{Strict R/U}, where a sample was deemed unrealistic if at least one clinician labeled it unrealistic. Responses to questions 2–3 were aggregated into mean clinician-perceived fidelity scores. Finally, we defined the \textbf{Deception Rate} as the fraction of synthetic cases judged to be real.

\begin{figure*}[h]
    \centering
    \subfigure[Original Data]{
        \centering
        \includegraphics[width=0.47\textwidth]{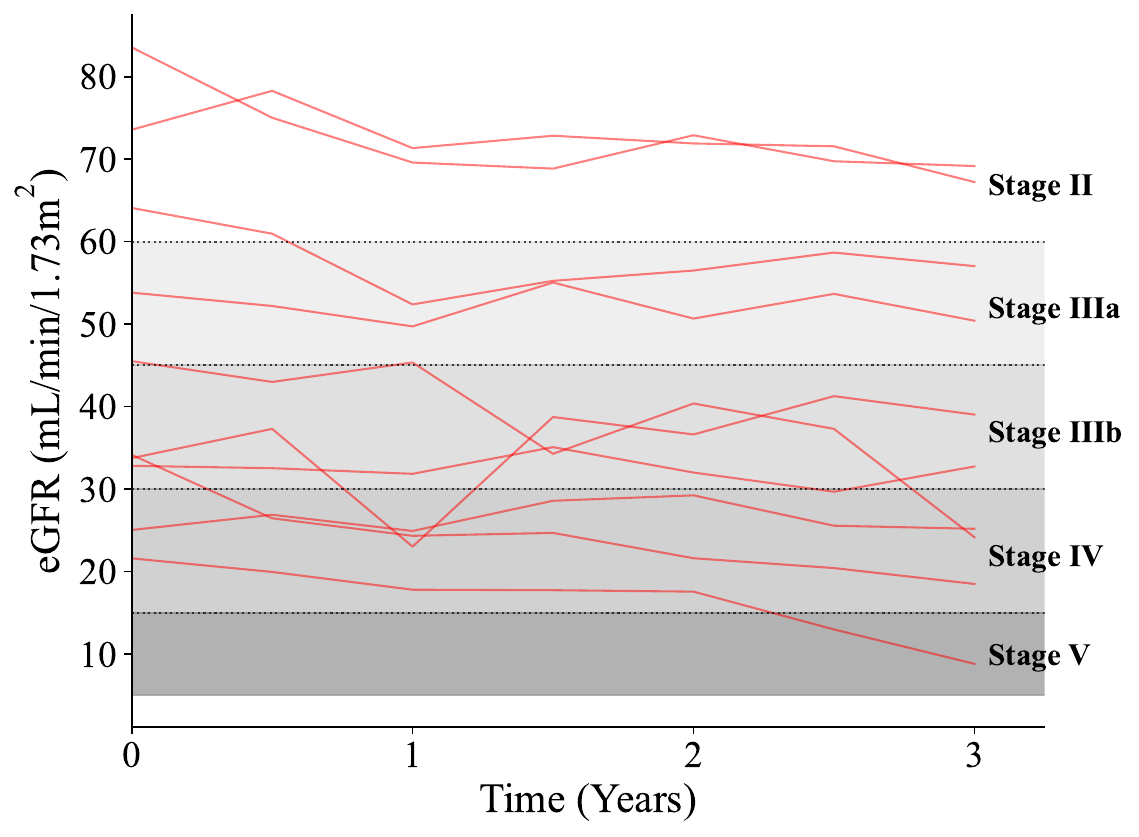}
        \label{fig:realGFR_paths}
        }
    \hfill
    \subfigure[TimeGAN]{
        \centering
        \includegraphics[width=0.47\textwidth]{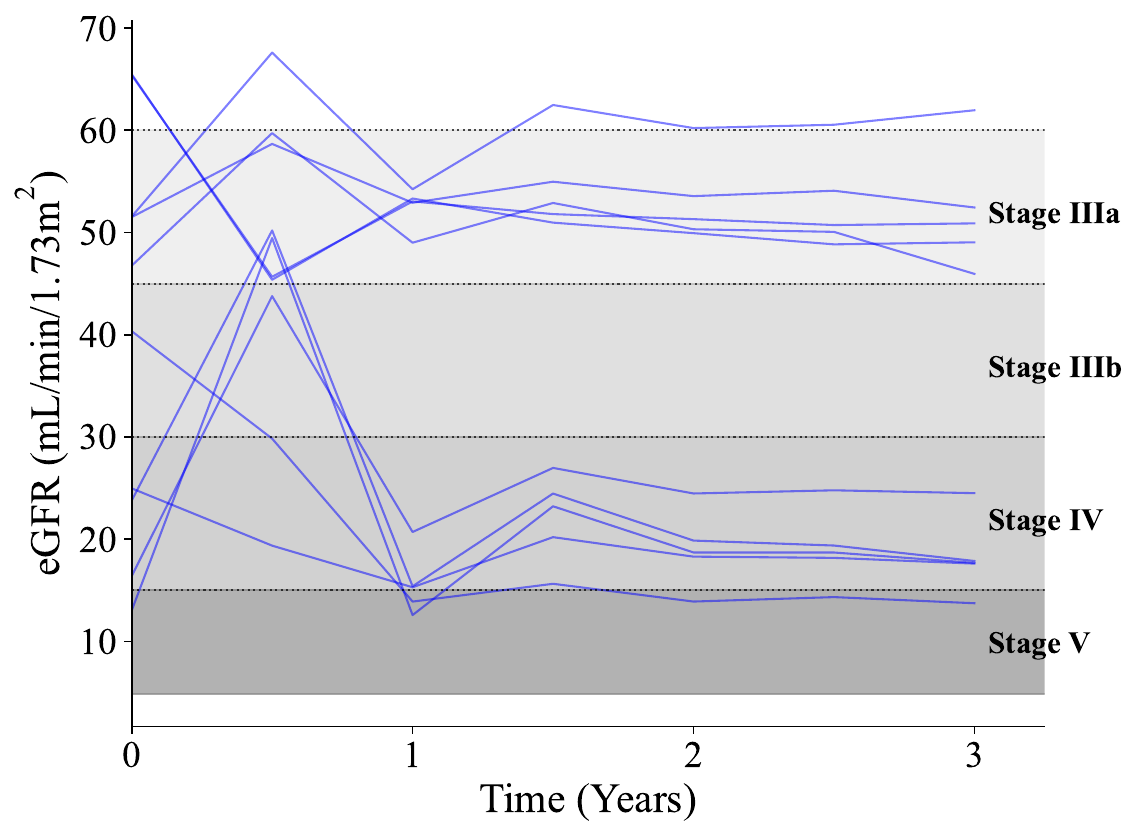}
        \label{fig:fakeGFR_paths}
        }
    \subfigure[Augmented TimeGAN]{
        \centering
        \includegraphics[width=0.47\textwidth]{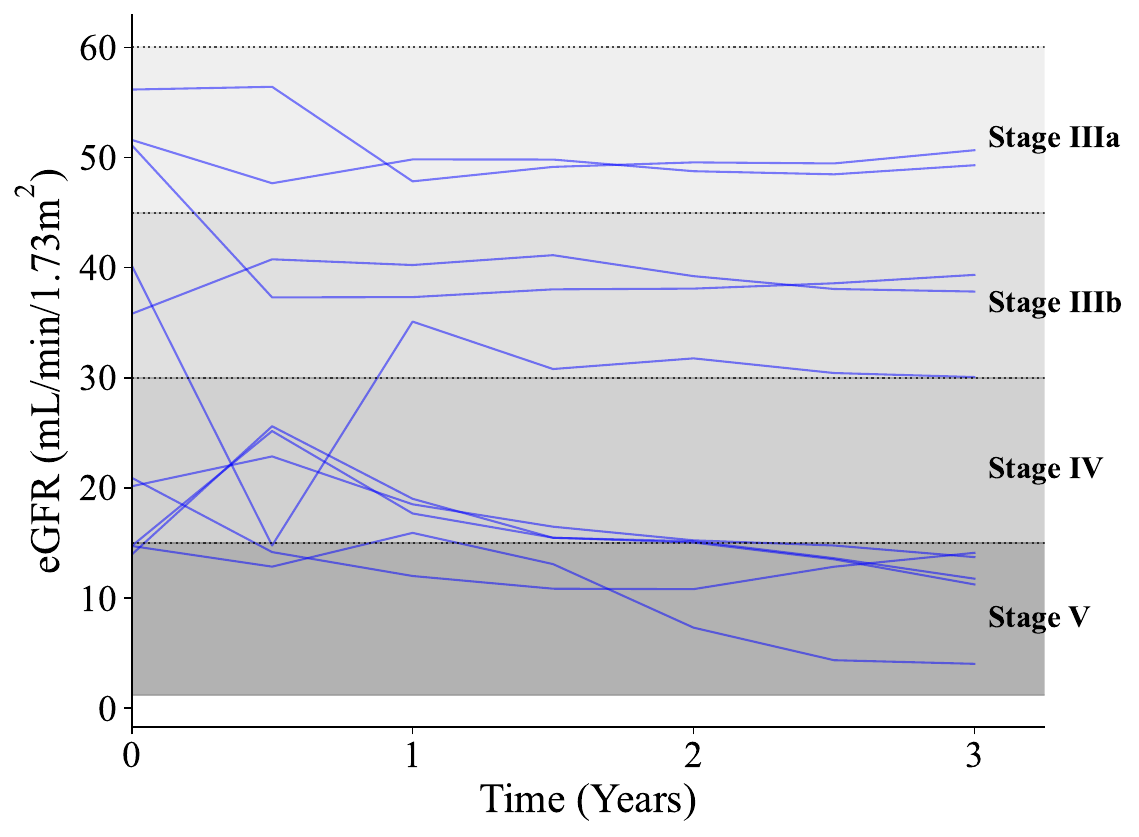}
        \label{fig:mod_fake_GFR_paths}
        }
    \hfill
    \subfigure[DP-TimeGAN]{
        \centering
        \includegraphics[width=0.47\textwidth]{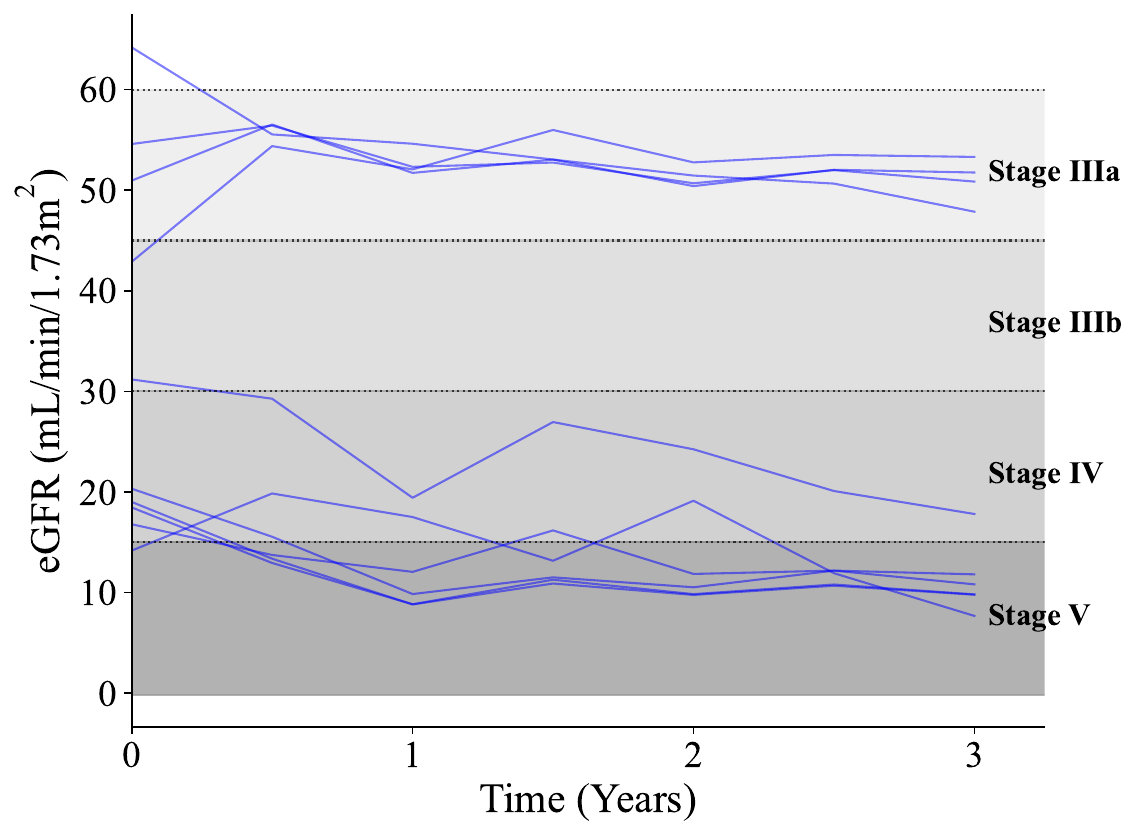}
        \label{fig:dp_fakeGFR_paths}
        }
    \caption[Real and synthetic eGFR trajectories for patients with chronic kidney disease.]{Real and synthetic eGFR trajectories for patients with chronic kidney disease (CKD). CKD stages are shaded in order of severity, labeled on the right. Data has shape ($N$, $T$, $C$) = (421, 7, 7); (b), (c), and (d) use parameters: $\# \text{epochs}= 10000$, $\# \text{layers} = 3$, $\text{latent-dim} = 24$, $\gamma=1$. For DP, ($\varepsilon$, $\delta$) = (10, $10^{-5}$).}
    \label{fig:egfr_synthesis}
\end{figure*}

\begin{table*}[t]
  \centering 
  \caption{Statistical metrics for synthetic data performance on the sines, eICU and CKD datasets using the following training parameters:  $\# \text{epochs}= 7000$, $\# \text{layers} = 3$, $\text{latent-dim} = 24$, $\text{noise-SD} = 0.2$. DP runs use $\varepsilon=15$ for sines, and $\varepsilon=20$ for eICU and CKD, alongside $\delta=10^{-5}$ for all runs. Benchmark models replicate the hyperparameters from their respective publications. All metrics are averaged over three training runs, and are listed as $\text{Mean} \pm \text{S.D.}$}
  \scalebox{0.8}{
  \begin{tabular}{>{\raggedright\arraybackslash}p{3.55cm} >{\raggedright\arraybackslash}p{2.25cm} >{\raggedright\arraybackslash}p{2.25cm} > {\raggedright\arraybackslash}p{2.55cm} > {\raggedright\arraybackslash}p{2.25cm} > {\raggedright\arraybackslash}p{2.85cm}}
  \toprule
    \textbf{Model} & \textbf{MMD ($\downarrow$)} & \textbf{DS ($\downarrow$)} & \textbf{$\alpha$-precision ($\uparrow$)} & \textbf{$\beta$-recall ($\uparrow$)} & \textbf{Authenticity ($\uparrow$)}\\
    \midrule
    \multicolumn{6}{l}{\textbf{Sines Dataset}} \\
    \midrule
    Augmented TimeGAN & $\mathbf{0.002 \pm 0.002}$ & $\mathbf{0.089 \pm 0.061}$ & $\mathbf{0.951 \pm 0.016}$ & $\mathbf{0.963 \pm 0.012}$ & $0.549 \pm 0.019$ \\ 
    DP-TimeGAN & $0.010 \pm 0.004$ & $0.213 \pm 0.056$ & $0.929 \pm 0.044$ & $0.918 \pm 0.022$ & $0.583 \pm 0.020$ \\
    SeriesGAN & $0.016 \pm 0.010$ & $0.203 \pm 0.104$ & $0.807 \pm 0.111$ & $0.799 \pm 0.086$ & $0.537 \pm 0.069$ \\
    DP Normalizing Flows & $0.020 \pm 0.011$ & $0.105 \pm 0.081$ & $0.602 \pm 0.106$ & $0.506 \pm 0.078$ & $\mathbf{0.598 \pm 0.114}$ \\
    TransFusion & $0.007 \pm 0.004$ & $0.257 \pm 0.070$ & $0.862 \pm 0.036$ & $0.865 \pm 0.031$ & $0.540 \pm 0.042$ \\
    TimeDiff & $0.018 \pm 0.010$  & $0.270 \pm 0.095$ & $0.657 \pm 0.134$ & $0.549 \pm 0.093$ & $0.576 \pm 0.103$\\
    \midrule
    \multicolumn{6}{l}{\textbf{eICU Dataset}} \\
    \midrule
    Augmented TimeGAN & $\mathbf{0.012 \pm 0.009}$ & $0.053 \pm 0.016$ & $\mathbf{0.951 \pm 0.038}$ & $0.941 \pm 0.032$ & $0.415 \pm 0.112$ \\ 
    DP-TimeGAN & $0.019 \pm 0.006$ & $0.145 \pm 0.072$ & $0.894 \pm 0.057$ & $0.920 \pm 0.029$ & $0.581 \pm 0.030$ \\
    SeriesGAN & $0.102 \pm 0.048$ & $0.240 \pm 0.061$ & $0.866 \pm 0.073$ & $0.788 \pm 0.040$ & $0.467 \pm 0.039$ \\
    DP Normalizing Flows & $0.020 \pm 0.013$ & $0.167 \pm 0.064$ & $0.776 \pm 0.079$ & $0.637 \pm 0.044$ & $\mathbf{0.684 \pm 0.068}$ \\
    TransFusion & $0.014 \pm 0.008$ & $\mathbf{0.032 \pm 0.012}$ & $0.942 \pm 0.027$ & $\mathbf{0.964 \pm 0.035}$ & $0.574 \pm 0.053$ \\
    TimeDiff & $0.018 \pm 0.009$ & $0.071 \pm 0.030$ & $0.698 \pm 0.104$ & $0.717 \pm 0.092$ & $0.532 \pm 0.135$\\
    \midrule
    \multicolumn{6}{l}{\textbf{CKD Dataset}} \\
    \midrule
    Augmented TimeGAN & $\mathbf{0.049 \pm 0.013}$ & $\mathbf{0.231 \pm 0.049}$ & $\mathbf{0.925 \pm 0.030}$ & $\mathbf{0.936 \pm 0.019}$ & $0.604 \pm 0.083$ \\ 
    DP-TimeGAN & $0.091 \pm 0.046$ & $0.312 \pm 0.053$ & $0.844 \pm 0.035$ & $0.904 \pm 0.047$ & $\mathbf{0.778 \pm 0.053}$ \\
    SeriesGAN & $0.125 \pm 0.062$ & $0.335 \pm 0.060$ & $0.880 \pm 0.021$ & $0.840 \pm 0.030$ & $0.569 \pm 0.104$ \\
    DP Normalizing Flows & $0.253 \pm 0.055$ & $0.323 \pm 0.071$ & $0.701 \pm 0.042$ & $0.718 \pm 0.041$ & $0.723 \pm 0.062$ \\
    TransFusion & $0.201 \pm 0.059$  & $0.344 \pm 0.082$ & $0.327 \pm 0.079$ & $0.488 \pm 0.092$ & $0.767 \pm 0.106$ \\
    TimeDiff & $0.073 \pm 0.040$ & $0.235 \pm 0.037$ & $0.494 \pm 0.065$ & $0.772 \pm 0.101$ & $0.703 \pm 0.086$ \\
    \bottomrule
  \end{tabular}
  }
  \label{tab:all_datasets_metrics} 
\end{table*}

\section{Clinical Datasets}
\label{datasets}

We evaluate generative performance on three datasets. As a benchmark, we use a synthetic sine dataset where sequences are generated as $\sin(\eta t + \theta)$ with $\eta \sim \mathcal{U}[0, 0.1]$ and $\theta \sim \mathcal{U}[0, 0.1]$ \citep{yoon_time-series_2019}. For clinical data, we first construct longitudinal EHRs from the eICU Collaborative Research Database, containing time-varying vital signs from intensive care unit (ICU) patients \citep{pollard_eicu_2017,pollard_eicu_2018}. Finally, to evaluate applicability in chronic disease, we use a chronic kidney disease (CKD) dataset with longitudinal estimated glomerular filtration rate (eGFR) trajectories, collected by \cite{iimori_data_2018,iimori_prognosis_2018}. Table~\ref{tab:feature_choices} details the shape and makeup of each dataset.

\subsection{Data Preprocessing} 
\label{datasets_preprocessing}

Each clinical dataset requires task-specific preprocessing to form consistent time-series tensor inputs for model training. For the eICU dataset, patient measurements ware resampled to a uniform interval of one observation per hour, and patients with incomplete or insufficient sequence lengths were removed. Remaining sequences were truncated or padded to a fixed length and reshaped into the standard RNN input format \citep{esteban_real-valued_2017}.

For the CKD dataset, longitudinal trajectories were constructed from patient time-series measurements, reshaped for RNN compatibility, and filtered to exclude incomplete sequences. Finally, both datasets were normalized using MinMax scaling.

\begin{table*}[t]
  \centering 
  \caption{Ablation study of the different unique portions of the Augmented and DP-TimeGAN on the sines dataset. TimeGAN models utilize the same parameters as mentioned in the caption of Table \ref{tab:all_datasets_metrics}. xLSTM-specific parameters include: $\# \text{heads} = 4$, $\# \text{blocks} = 4$, $\text{sLSTM positions} = 1, 3$, $\text{1D convolution kernel size} = 4$, $\text{QKV block size} = 4$, $\text{projection factor} = 1.3$ and $\text{activation function} = \text{GeLU}$. All metrics are calculated from three separated training runs, and are listed as $\text{Mean} \pm \text{S.D.}$}
  \scalebox{0.8}{\begin{tabular}{>{\raggedright\arraybackslash}p{4cm} >{\raggedright\arraybackslash}p{2.25cm} >{\raggedright\arraybackslash}p{2.25cm} > {\raggedright\arraybackslash}p{2.25cm} > {\raggedright\arraybackslash}p{2.25cm} > {\raggedright\arraybackslash}p{2.25cm}}
  \toprule
    \textbf{Modifications} & \textbf{MMD} & \textbf{DS} & \textbf{$\alpha$-precision} & \textbf{$\beta$-recall} & \textbf{Authenticity}\\
    \midrule
    \multicolumn{6}{l}{\textbf{Augmented TimeGAN}} \\
    \midrule
    None & $0.008 \pm 0.004$ & $0.269 \pm 0.044$ & $0.648 \pm 0.157$ & $0.657 \pm 0.157$ & $0.531 \pm 0.040$ \\
    xLSTM & $0.012 \pm 0.011$ & $0.289 \pm 0.042$ & $0.549 \pm 0.279$ & $0.546 \pm 0.223$ & $0.451 \pm 0.095$ \\ 
    \textbf{Noise Injection} & $\mathbf{0.002 \pm 0.002}$ & $\mathbf{0.089 \pm 0.061}$ & $\mathbf{0.951 \pm 0.016}$ & $\mathbf{0.963 \pm 0.012}$ & $\mathbf{0.549 \pm 0.019}$ \\
    xLSTM \& Noise Injection & $0.009 \pm 0.011$ & $0.290 \pm 0.053$ & $0.856 \pm 0.025$ & $0.916 \pm 0.049$ & $0.495 \pm 0.119$  \\
    \midrule
    \multicolumn{6}{l}{\textbf{DP-TimeGAN}} \\
    \midrule
    None & $0.015 \pm 0.010$ & $0.237 \pm 0.154$ & $0.664 \pm 0.233$ & $0.837 \pm 0.096$ & $\mathbf{0.682 \pm 0.124}$ \\
    xLSTM & $0.012 \pm 0.002$ & $0.304 \pm 0.088$ & $0.897 \pm 0.044$ & $0.837 \pm 0.074$ & $0.546 \pm 0.128$ \\
    \textbf{Noise Injection} & $\mathbf{0.010 \pm 0.004}$ & $\mathbf{0.213 \pm 0.056}$ & $\mathbf{0.929 \pm 0.044}$ & $\mathbf{0.918 \pm 0.022}$ & $0.583 \pm 0.020$ \\
    xLSTM \& Noise Injection & $0.020 \pm 0.010$ & $0.257 \pm 0.056$ & $0.704 \pm 0.195$ & $0.817 \pm 0.213$ & $0.629 \pm 0.034$ \\
    \bottomrule
  \end{tabular}
  }
  \label{tab:ablation}
\end{table*}

\section{Results} 

\label{results}

Figure \ref{fig:egfr_synthesis} shows a sampling of the real and resultant synthetic CKD progression eGFR pathways from the TimeGAN, Augmented TimeGAN, and DP-TimeGAN models. We observe that the synthetic sequences capture transitions between disease stages, even in relatively early progression of CKD, where training data is scarce. Of the three models, the Augmented TimeGAN displays the strongest preservation of the original data distribution, while the original TimeGAN model struggles to capture transitions between CKD stages. Additionally, Figure \appendixref{fig:dp_fakeGFR_paths} validates the notion that the differentially private variant of the model sacrifices fidelity and diversity as compared to the augmented TimeGAN model. To quantify model performance and practicality in a clinical context, we use both statistical and clinician-evaluated measures.

\subsection{Statistical Performance Measures} 
\label{results_statistical_measures}

We first evaluate synthetic samples using the statistical metrics described in Sections~\ref{eval_fidelity_diversity_privacy} and~\ref{eval_downstream_utility}, enabling reproducible benchmarking without requiring clinician input. Results are reported in Tables~\ref{tab:all_datasets_metrics} and~\ref{tab:utility}.

DP-TimeGAN achieves the strongest authenticity scores on both the sine and CKD datasets but is surpassed by DP Normalizing Flows on eICU authenticity, though at the cost of reduced sample quality. Conversely, DP-TimeGAN underperforms in fidelity and diversity, where Augmented TimeGAN demonstrates superior results, which demonstrates the fidelity tradeoff of differentially private training \citep{esteban_real-valued_2017}. TransFusion performs comparably to Augmented TimeGAN on the eICU dataset but suffers on others. While DP Normalizing Flows outperforms Augmented TimeGAN on downstream AUC-ROC, it fails to produce a strong Predictive Score, limiting the utility of outputs for longitudinal prediction tasks. These quantitative findings only build a partial picture, and motivate further validation through blinded clinician review.

\begin{table*}
  \centering 
  \caption{Clinician validation of 25 random patient longitudinal EHRs from the real dataset and the models. The TimeGAN models utilize the same parameters as mentioned in Table \ref{tab:all_datasets_metrics}.}
  \scalebox{0.9}{
  \begin{tabular}{>{\raggedright\arraybackslash}p{3.6cm} >{\centering\arraybackslash}p{2.2cm} >{\centering\arraybackslash}p{2.1cm} >{\centering\arraybackslash}p{1.8cm} >{\centering\arraybackslash}p{1.8cm} >{\centering\arraybackslash}p{2.0cm}}
  \toprule
    \textbf{Model} & \textbf{Relaxed R/U} & \textbf{Strict R/U} & \textbf{Q2 Mean} & \textbf{Q3 Mean} & \textbf{Deception Rate}\\
    \midrule
    Real Data & 0.857 & 0.143 & 3.286 & 3.036 & - \\
    Regular TimeGAN & 1.000 & 0.250 & 3.813 & 3.438 & 0.750 \\
    Augmeted TimeGAN & 1.000 & 0.800 & 3.700 & 3.250 & \textbf{0.960} \\ 
    DP-TimeGAN & 1.000 & 1.000 & 4.354 & 4.229 & 0.950 \\
    \bottomrule
  \end{tabular}
  }
  \label{tab:clinical_validation} 
\end{table*}

\begin{table}[t]
  \centering 
  \caption{Utility metrics for generative models trained on the CKD dataset. TimeGAN-based models use the parameters shown in Table \ref{tab:all_datasets_metrics}. Benchmark models replicate the parameters from their respective publications. All metrics are averaged over three training runs, and are listed as $\text{Mean} \pm \text{S.D.}$}
  \scalebox{0.8}{
  \begin{tabular}{>{\raggedright\arraybackslash}p{3.5cm} >{\raggedright\arraybackslash}p{2cm} >{\raggedright\arraybackslash}p{2cm}}
  \toprule
    \textbf{Model} & \textbf{Predictive Score} & \textbf{Downstream AUC-ROC} \\
    \midrule
    Real Data & $0.289 \pm 0.016$ & $0.730 \pm 0.117$ \\
    Augmented TimeGAN & $0.381 \pm 0.050$ & $0.615 \pm 0.046$  \\ 
    DP-TimeGAN & $\mathbf{0.370 \pm 0.033}$ & $0.549 \pm 0.063$  \\
    SeriesGAN & $0.448 \pm 0.052$ & $0.535 \pm 0.050$ \\
    DP Normalizing Flows & $0.497 \pm 0.062$ & $\mathbf{0.648 \pm 0.008}$ \\ 
    TransFusion & $0.395 \pm 0.051$ & $0.450 \pm 0.104$ \\
    TimeDiff & $0.565 \pm 0.122$ & $0.540 \pm 0.140$ \\
    \bottomrule
  \end{tabular}
  }
  \label{tab:utility} 
\end{table}

\subsection{Clinical Validation} 
\label{results_clinical_validation}

We complement statistical metrics by conducting blinded expert clinician evaluations of CKD trajectories (see Section~\ref{eval_clinician_evaluation}).
Using the Relaxed R/U criterion, 86\% of real patients and 100\% of synthetic patients were judged realistic. Under the Strict R/U criterion, 14\% of real patients and 50\% of synthetic patients were labeled realistic. We learn that high-fidelity synthetic records at are practically imperceptable at a high level, even to a subset of expert clinicians.
A detailed breakdown by model is provided in Table~\ref{tab:clinical_validation}. Overall, results suggest that synthetic data achieves parity with real CKD trajectories in the most conservative evaluation setting.

\section{Discussion}
\label{discussion}

DP-TimeGAN provides a secure framework for generating synthetic longitudinal EHRs, offering stronger privacy protection than baseline models, as evidenced by improved authenticity scores. The model also produces clinically useful data, as demonstrated by its AUC-ROC performance on diabetes prediction in CKD patients and by clinician assessments rating its trajectories as realistic. To our knowledge, this is the first approach to combine formal privacy guarantees with demonstrated clinical realism and downstream utility in chronic disease EHRs.

The choice to incorporate formal privacy-preserving mechanisms supports compliance with HIPAA and GDPR, enabling a safer integration of ML tools into clinical workflows. Furthermore, we see this strategy as an opportunity to expand data accessibility for CKD research and pave the way toward broader applications in chronic disease modeling.

\paragraph{Limitations}
\label{limitations}

We acknowledge several limitations that should be overcome to generalize our results into broader applicability. First, the CKD dataset contains a limited set of features measured over a fixed three-year window, where only eGFR is longitudinal, while all other measurements are held static at the initial observation. This constrains the diversity and expressivity of CKD trajectories available for training, and was highlighted by clinicians as a drawback during expert classification.

Moreover, we draw attention to the inherent privacy–utility trade-off introduced with differential privacy. As shown in Appendix~\ref{appendix_privacy_utility}, DP-TimeGAN sacrifices some data quality to enforce quantifiable privacy guarantees, which may reduce its effectiveness for downstream tasks. The choice of privacy budget remains underexplored, and further work is needed to systematically characterize trade-offs among privacy, utility, and computational cost.

Finally, we pose as future work that model performance may be improved by incorporating alternative DP mechanisms, architectural refinements such as state-space model (SSM) backbones, or training stabilization techniques including spectral normalization or a Wasserstein loss \citep{gulrajani_improved_2017,miyato_spectral_2018, zhang_effectively_2023}.

\section{Conclusion}
We introduce DP-TimeGAN, a privacy-preserving generative model for synthesizing longitudinal electronic health records (EHRs) in chronic disease contexts. DP-TimeGAN demonstrates stronger formal privacy guarantees than baseline models while maintaining near-state-of-the-art performance on downstream predictive tasks and blinded clinical evaluations for CKD. Clinicians rate both Augmented TimeGAN and DP-TimeGAN outputs as clinically realistic, aligning with patterns observed in practice. 

When combined with disease progression models, DP-TimeGAN has direct applications in disease modeling research, mitigating barriers to longitudinal EHR access, accelerating medical software testing, and informing healthcare delivery economics.

\acks{We gratefully acknowledge the essential contributions of Vironix Health, colleagues at Stanford University, and the Nephrology Division at Unity Health Toronto, whose collective expertise provided the clinical foundation and expert data evaluation within this paper. \\

We thank the following clinicians for providing the review discussed in Section \ref{results_clinical_validation}: Dr. Indira Chevru (Georgia Kidney Associates, Marietta, GA), Dr. Vartan Papazian (Modern Nephrology, Antelope Valley, CA), and Dr. Nicholas Wysham (Vancouver Clinic, Vancouver, WA).}

\bibliography{bibliography}

\appendix

\begin{table*}
    \centering
    \caption{Feature choices for datasets being used in DP-TimeGAN evaluation.}
    \scalebox{0.9}{
    \begin{tabular}{>{\raggedright\arraybackslash}p{4cm} >{\raggedright\arraybackslash}p{8cm} >{\raggedright\arraybackslash}p{3cm}}
        \toprule
        \textbf{Dataset} & \textbf{Feature Choices} & \textbf{Dataset shape}$^*$\\
        \midrule
        \textbf{Sines dataset} \citep{yoon_time-series_2019} & Synthetically generated sine waves & (500, 24, 5) \\
        \cmidrule(lr){1-3}
        \textbf{eICU Collaborative Research Database} \citep{pollard_eicu_2017} & Body temperature, oxygen saturation, heart rate, mean blood pressure, respiration rate & (750, 12, 5) \\
        \midrule
        \textbf{Chronic kidney disease dataset} \citep{iimori_data_2018} & Age, body mass index, estimated glomerular filtration rate (eGFR), albumin, hemoglobin, creatinine, urine protein-to-creatinine ratio & (421, 7, 7) \\
        \cmidrule(lr){1-3}
        \cmidrule(lr){1-3}
    \end{tabular}
    }
    \label{tab:feature_choices}
    \par\raggedright
    {\footnotesize \quad \qquad \: *\textit{Shapes are taken after cleaning.}}
\end{table*}

\begin{figure*}[htp]
    \centering  
    \subfigure[Original Data]{
        \centering
        \includegraphics[width=0.48\textwidth]{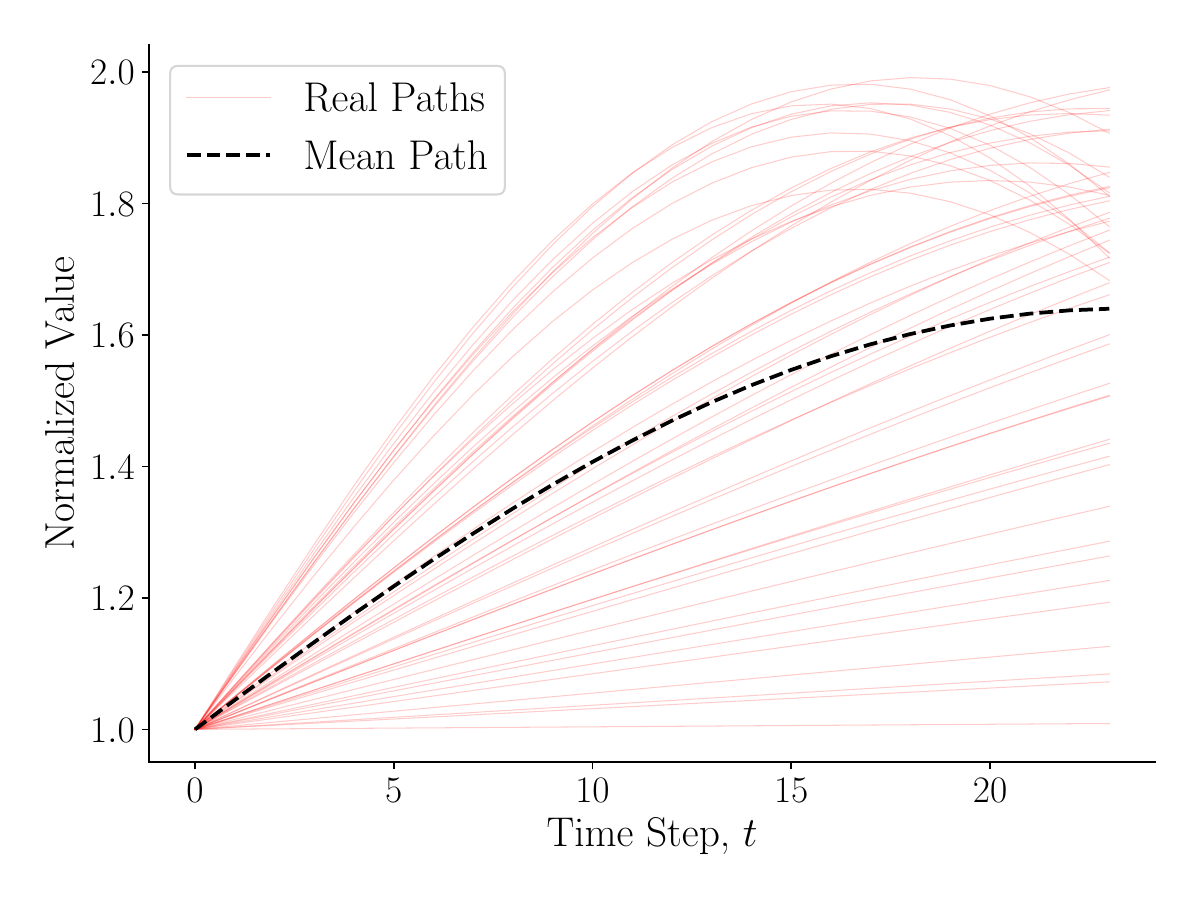}
        \label{fig:realsine_paths}
        }
    \subfigure[Synthetic Data]{
        \centering
        \includegraphics[width=0.48\textwidth]{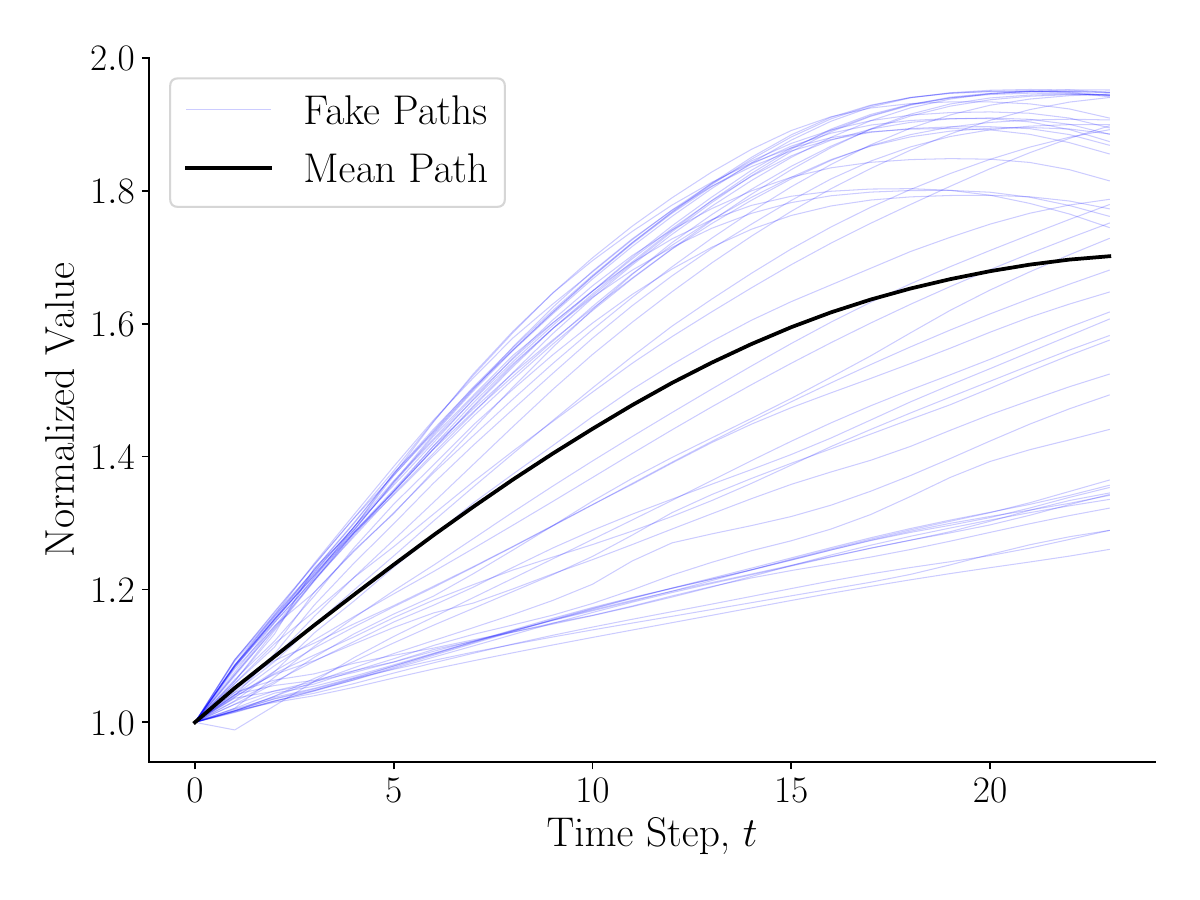}
        \label{fig:fakesine_paths}
        }
    \vspace{0.4em}
    \subfigure[PCA visualization]{
        \centering
        \includegraphics[width=0.48\textwidth]{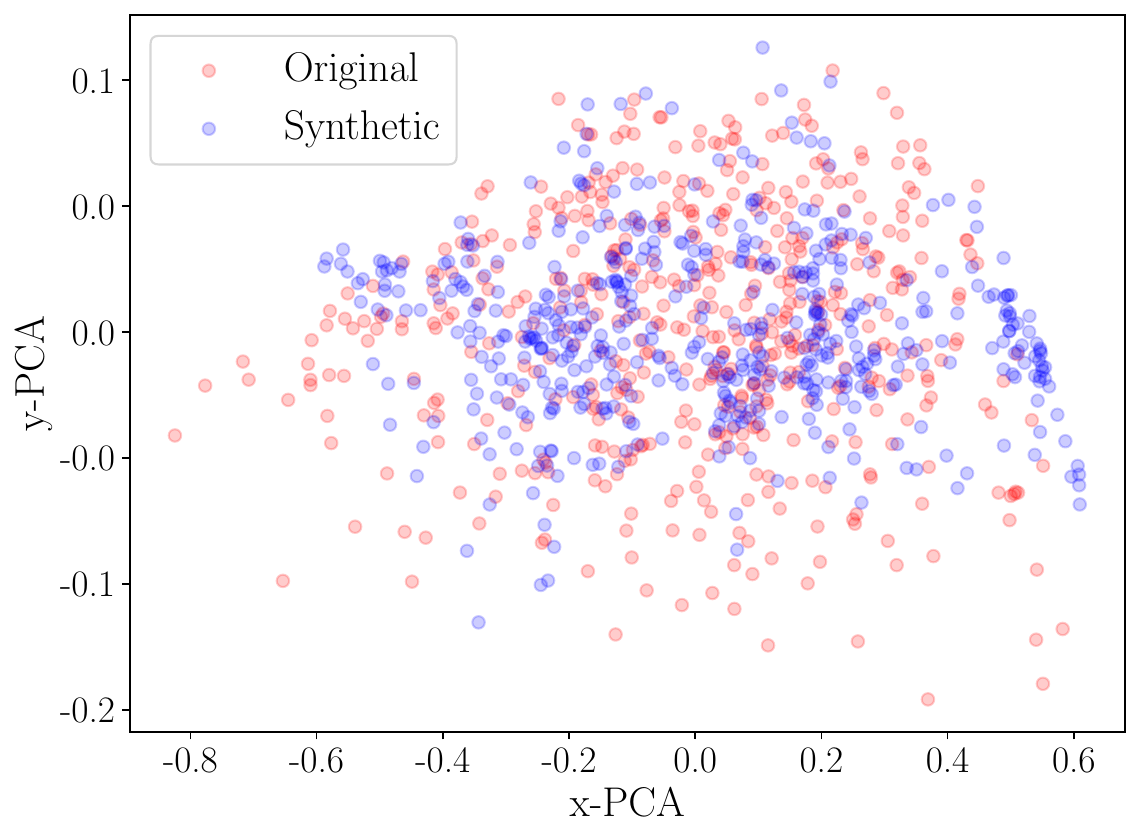}
        \label{fig:pca_sines}
        }
    \subfigure[t-SNE visualization]{
        \centering
        \includegraphics[width=0.48\textwidth]{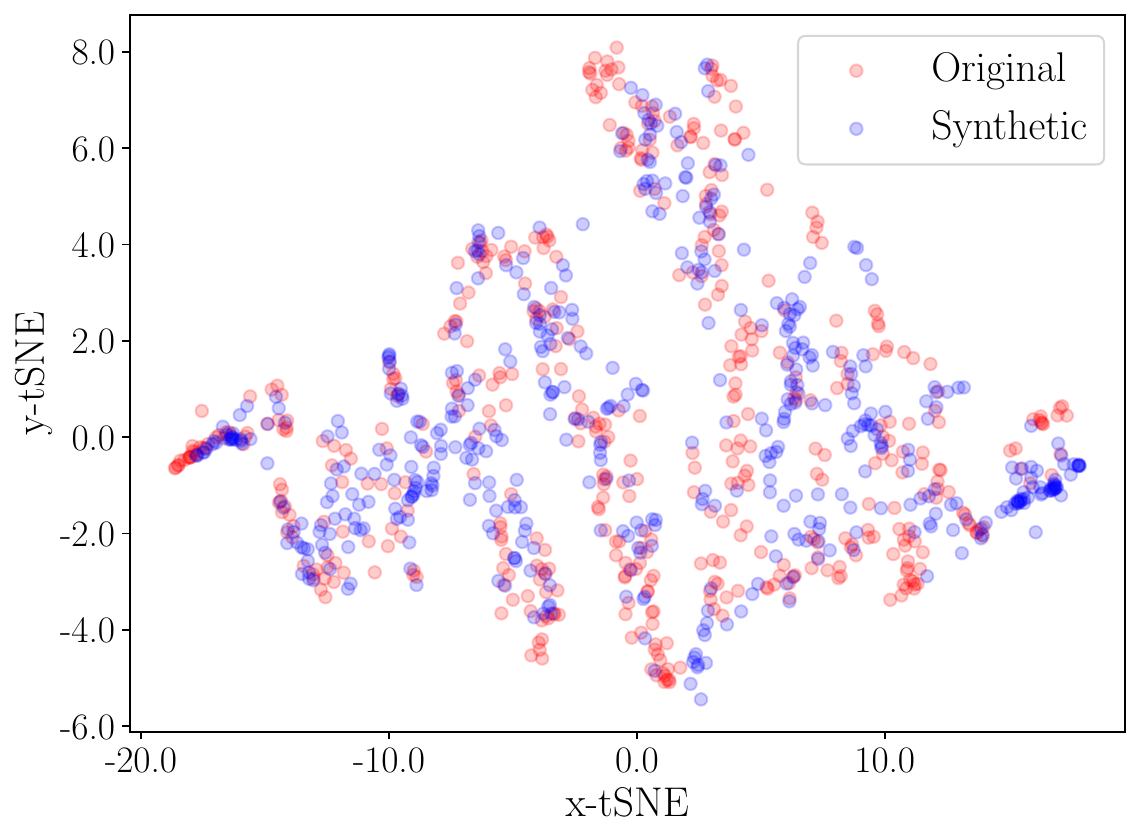}
        \label{fig:tsne_sines}
        }
    \caption[Comparison of real and synthetic sinusoidal data]{Comparison of real and synthetic sinusoidal data from the Augmented TimeGAN. Real data uses ($N$, $T$, $C$) = (700, 24, 5), where each feature is a randomly generated sine wave; training parameters are: $\# \text{epochs} = 6000$, $\# \text{layers} = 3$ and $\text{latent-dim} = 24$. All data is normalized to a starting value of 1 prior to plotting for clarity. Plots (a) and (b) isolate one feature of real and synthetic sine waves, respectively; plots (c) and (d) compare the PCA and t-SNE results, respectively, for the two datasets.}
    \label{fig:aug_sine_synthesis}
\end{figure*}

\section{Augmented TimeGAN Ablation Study}
\label{appendix_ablation}

To measure the effects of each modification introduced into the Augmented TimeGAN, we performed an ablation study with the sines dataset, as shown in Table \ref{tab:ablation}. Figure \ref{fig:aug_sine_synthesis} also illustrates the capabilities of Augmented TimeGAN on the sine dataset. The Augmented TimeGAN version that had the best performance was including the noise injection only. Furthermore, the Augmented TimeGAN outperformed the original TimeGAN on the eICU and CKD datasets, as shown in Tables \ref{tab:metrics_reg_timegan} and \ref{tab:utility_reg_timegan}. Therefore, when creating DP-TimeGAN, we chose to only integrate the noise injection for the best possible performance before integrating differential privacy. 

\begin{table*}[t]
  \centering 
  \caption{Regular and Augmented TimeGAN statistical metrics for synthetic data performance on the eICU and CKD datasets. TimeGAN models utilize the same parameters as mentioned in Table \ref{tab:all_datasets_metrics}. All metrics are calculated from three separated training runs, and are listed as $\text{Mean} \pm \text{S.D.}$}
  \scalebox{0.9}{\begin{tabular}{>{\raggedright\arraybackslash}p{4cm} >{\raggedright\arraybackslash}p{2.25cm} >{\raggedright\arraybackslash}p{2.25cm} > {\raggedright\arraybackslash}p{2.25cm} > {\raggedright\arraybackslash}p{2.25cm} > {\raggedright\arraybackslash}p{2.25cm}}
  \toprule
    \textbf{Model} &  \textbf{MMD} & \textbf{DS} & \textbf{$\alpha$-precision} & \textbf{$\beta$-recall} & \textbf{Authenticity}\\
    \midrule
    \multicolumn{6}{l}{\textbf{eICU Dataset}} \\
    \midrule
    Original TimeGAN & $0.161 \pm 0.130$ & $0.153 \pm 0.096$ & $0.792 \pm 0.078$ & $0.746 \pm 0.016$ & $\mathbf{0.446 \pm 0.104}$ \\
    Augmented TimeGAN & $\mathbf{0.012 \pm 0.009}$ & $\mathbf{0.053 \pm 0.016}$ & $\mathbf{0.951 \pm 0.038}$ & $\mathbf{0.941 \pm 0.032}$ & $0.415 \pm 0.112$ \\
    \midrule
    \multicolumn{6}{l}{\textbf{CKD Dataset}} \\
    \midrule
    Original TimeGAN & $0.061 \pm 0.015$ & $0.341 \pm 0.112$ & $0.848 \pm 0.028$ & $0.841 \pm 0.020$ & $\mathbf{0.613 \pm 0.083}$ \\
    Augmented TimeGAN & $\mathbf{0.049 \pm 0.013}$ & $\mathbf{0.231 \pm 0.049}$ & $\mathbf{0.925 \pm 0.030}$ & $\mathbf{0.936 \pm 0.019}$ & $0.604 \pm 0.083$ \\  
    \bottomrule
  \end{tabular}
  }
  \label{tab:metrics_reg_timegan} 
\end{table*}

\begin{table}[t]
  \centering 
  \caption{Utility metrics for Regular and Augmented TimeGAN trained on the CKD dataset. TimeGAN models utilize the same parameters as mentioned in Table \ref{tab:all_datasets_metrics}. All metrics are calculated from three separated training runs, and are listed as $\text{Mean} \pm \text{S.D.}$}
  \scalebox{0.9}{
  \begin{tabular}{>{\raggedright\arraybackslash}p{3cm} >{\raggedright\arraybackslash}p{2cm} >{\raggedright\arraybackslash}p{2cm}}
  \toprule
    \textbf{Model} & \textbf{Predictive Score} & \textbf{Downstream AUC-ROC} \\
    \midrule
    Original TimeGAN & $0.443 \pm 0.048$ & $0.564 \pm 0.052$ \\[1em]
    Augmented TimeGAN & $\mathbf{0.381 \pm 0.050}$ & $\mathbf{0.615 \pm 0.046}$  \\ 
    \bottomrule
  \end{tabular}
  }
  \label{tab:utility_reg_timegan} 
\end{table}

\section{Differential Privacy in Data Releases}

To justify our $(\epsilon, \delta)$ choices when training DP-TimeGAN, we refer to Table \ref{tab:epsilon_values}. Maintaining $\epsilon \in [10, 20]$ and $\delta = 10^{-5}$ fits within the various data releases conducted in governmental and private agencies, meaning that they are acceptable values for DP-TimeGAN experiments. These data releases are also the best precedent for DP parameters that would be legally allowed, as the best method to choose $\epsilon$ is not clear and is still an open question based on government reports. Furthermore, a typical requirement is that $\delta < \frac{1}{n}$, for a dataset containing $n$ patients. This is maintained as per the values in Table \ref{tab:feature_choices} \citep{near_guidelines_2025}.

\begin{table*}[t]
    \centering
    \caption{$(\varepsilon, \delta)$ values for differential privacy in various data releases.}
    \scalebox{0.9}{
    \begin{tabular}{>{\raggedright\arraybackslash}p{5.5cm} >{\raggedright\arraybackslash}p{5cm} >{\raggedright\arraybackslash}p{4cm}}
        \toprule
        \textbf{Use Case} & \textbf{Data Released} & \textbf{ ($\varepsilon$, $\delta$) Values} \\
        \midrule
        \textbf{US Census Bureau} (2020 Census) \citep{uscensus2020keyparameters} & Population data, demographic characteristics & (19.6, 1e-5) \\
        \cmidrule(lr){1-3}
        \textbf{Meta} (Pandemic Motility) \citep{facebook2020mobilitydata} & User motility data during COVID-19 pandemic & (2.0, 0) \\
        \cmidrule(lr){1-3}
        \textbf{Apple} (QuickType) \citep{apple2021privacy} & User vocabulary on iOS keyboards & (4.0, 0) \\
        \cmidrule(lr){1-3}
        \textbf{LinkedIn} (Audience Engagement) \citep{rogers2020linkedin} & User activity and content engagement trends & (0.15, 1e-10) \\
        \bottomrule
    \end{tabular}
    }
    \label{tab:epsilon_values}
\end{table*}

\section{Metric Calculation Details}
\label{appendix_metric}

\subsection{Fidelity Metrics}
\label{appendix_fidelity_metrics}
For fidelity metrics, we consider the (i) maximum mean discrepancy, (ii) discriminative score, and (iii) $\alpha$-precision, each summarized below.

\vspace{1ex}

\noindent \textbf{Maximum mean discrepancy (MMD)} is a multivariate distributional distance metric that transforms data using the Gaussian kernel:

\begin{equation*}
K(x, y) = \exp\qty{-\frac{\|x-y\|^2}{2\sigma^2}}
\end{equation*}

to compute a distance between two distributions. Here, $\sigma$ is a length scale parameter that we take to equal 1 in our tests. For real and synthetic datasets, $\vb{x}$ and $\vb{\hat{x}}$ consisting of $N$ and $M$ sequences, respectively, we first obtain cross-sectional latent codes.

\begin{align*}
\vb{h}^{(n)} &= \mathcal{E}(\vb{x}^{(n)}) \qquad \text{for } n = 1, \ldots, N, \\
\vb{\hat{h}}^{(n)} &= \mathcal{E}(\vb{\hat{x}}^{(n)}), \qquad \text{for } n = 1, \ldots, M.
\end{align*}

Using the cross-sectional data from the above two equations, the MMD is calculated as:

\begin{equation*}
\begin{split}
\text{MMD}^2(\vb{h}, \vb{\hat{h}}) = & \; \mathbb{E}_{\vb{h},\vb{h^{\prime}}}\left[K(\vb{h}, \vb{h^{\prime}})\right] \\ 
& +  \mathbb{E}_{\vb{\hat{h}},\vb{\hat{h}^{\prime}}}\left[K(\vb{\hat{h}}, \vb{\hat{h}^{\prime}})\right] \\
& -2\mathbb{E}_{\vb{h},\vb{\hat{h}}}\left[K(\vb{h}, \vb{\hat{h}})\right]
\end{split}
\end{equation*}

where abstractly, $\mathbb{E}_{\vb{u},\vb{v}}[K(\vb{u}, \vb{v})]$ represents the expectation of the Gaussian kernel $K$ over all vector pairs $(\vb{u}, \vb{v})$ sampled from the respective datasets; $\vb{h^{\prime}}$ and $\vb{\hat{h}^{\prime}}$ are used when the comparison is occuring within the same dataset. We choose this metric because it is directly applicable to multivariate distributions, and does not require kernel density estimation, enabling superior reproducibility.

\vspace{1ex}

\noindent \textbf{Discriminative score} trains a separate recurrent neural network to classify real and synthetic sequences. The training dataset is composed of a sample of entries from the real and synthetic data which have been labeled according to their validity. After the classifier has been trained for a fixed number of epochs, the model is tested on an unseen sample of data, and the discriminative score is calculated as

\begin{equation*}
    \mathrm{DS} = \left| 0.5 - \frac{N_{correct}}{N_{total}}\right|
\end{equation*}

Where $N_{total}$ is the total number of sequences in the testing dataset, and $N_{correct}$
is the number of sequences correctly classified by the recurrent classifier. Intuitively, discriminative score measures the extent to which the classifier’s accuracy matches that of a random guess (lower is better from perspective of generated data fidelity). For additional details on this metric, please see \cite{yoon_time-series_2019}.

\vspace{1ex}

\noindent$\mathbf{\alpha}$\textbf{-precision} measures the probability that a sample from the synthetic data resides within the $\alpha$-support of the real data distribution. 

\begin{equation*}
P_{\alpha} \triangleq \mathbb{P}(\vb{\hat{x}}^{(i)} \in \mathcal{S}_r^{\alpha}), \text{ for } \alpha \in [0, 1].   
\end{equation*}

where $\mathcal{S}_r$ is the distribution of the real data.
For additional details on this metric, please see \cite{alaa_how_2022}.

\subsection{Diversity Metrics}
\label{appendix_diversity_metrics}

\noindent \textbf{Principal Component Analysis (PCA)} is a synthetic data visualization tool that is produced by first centering each real and  synthetic sequences about their temporal means, then aggregating the centered vectors, $\mathbf{c}_t \in \mathbb{R}^n$, into a matrix, $\mathbf{C} \in \mathbb{R}^{T \times n}$:
\begin{align}
\mathbf{c}_t &= \mathbf{x}_t - \frac{1}{T} \sum_{t=1}^{T} \mathbf{x}_t, \quad \text{for } t = 1,\ldots,T. \label{eqn:center xt} \\
\mathbf{C} &= [\mathbf{c}_1, \mathbf{c}_2, \ldots, \mathbf{c}_T]^\top,
\label{eqn: matrix c}
\end{align}

Finally computing a singular value decomposition of $\mathbf{C}$, we may use the first two right singular vectors to project the centralized data:
\begin{align}
    \mathbf{C} &= \mathbf{U}\boldsymbol{\Sigma}\boldsymbol{V}^\top, \\
    \mathbf{C}_{\text{PCA}} &= \mathbf{C}\begin{bmatrix}
\mathbf{v}_1 \,|\, \mathbf{v}_2
\end{bmatrix}.
\label{eqn: final pca}
\end{align}

Here, $\mathbf{v}_1$, and $\mathbf{v}_2$ are the first two right singular vectors, extracted from $\mathbf{V}$. The result, $\mathbf{C}_{\text{PCA}} \in \mathbb{R}^{n \times 2}$, has rows which indicate (x, y) coordinates in the projected space, that may be used for 2D visualization. We perform \eqref{eqn:center xt}-\eqref{eqn: final pca} to each sequence in $\{\mathbf{x}^{(i)}_{1:T}\}_{i=1}^N$ and $\{\hat{\mathbf{x}}^{(i)}_{1:T}\}_{i=1}^M$, visualizing both results to highlight whether the span and clusters in the datasets are in alignment, from which we draw insights about diversity.

\vspace{1ex}

\noindent \textbf{t-Distributed Stochastic Neighbor Embedding (t-SNE)}  is a non-linear dimensionality reduction technique that works by converting pairwise Euclidean distances between high-dimensional points into conditional probabilities representing similarities. The algorithm then finds a lower dimensional embedding that best preserves these similarities using a gradient descent method. The result in our case is a two-dimensional map where points that were nearby in the original high-dimensional space remain close, making it an effective tool for visualizing complex local structures such as clusters in synthetic and real datasets. For additional details on this metric, including the mathematical formulation, please see \cite{JMLR:v9:vandermaaten08a}.

\vspace{1ex}

\noindent \textbf{$\beta$-recall} measures the probability that a sample from the real data resides within the $\beta$-support of the synthetic data distribution.

\begin{equation*}
R_{\beta} \triangleq \mathbb{P}(\vb{{x}}^{(i)} \in \mathcal{S}_g^{\beta}), \text{ for } \beta \in [0, 1].   
\end{equation*}

where $\mathcal{S}_g$ is the distribution of the synthetic data.
For additional details on this metric, please see \cite{alaa_how_2022}.

\subsection{Privacy Metric}
\label{appendix_privacy_metric}

\textbf{Authenticity} measures the probability of a generative model synthesizing unique samples rather than copies of training data that are slightly shifted.

\begin{equation*}
\mathbb{P}_g = A \cdot \mathbb{P}'_{g} + (1-A) \cdot \delta_{g, \epsilon}
\end{equation*}

where $\delta_{g, e} = \delta_{g} \ast \mathcal{N}(0, \epsilon^2)$ and $\delta_g$ is a specified probability mass function for the training data. For additional details on this metric, please see \cite{alaa_how_2022}.

\subsection{Utility Metrics}
\label{appendix_utility_metrics}

\begin{table}[t]
  \centering 
  \caption{Hyperparameters for the downstream GRU classifier}. 
  \scalebox{0.9}{
  \begin{tabular}{>{\raggedright\arraybackslash}p{3cm} > {\raggedright\arraybackslash}p{2.5cm} }
  \toprule
    \textbf{Hyperparameter} & \textbf{Value} \\
    \midrule
    Data normalization & Standard scaler \\
    Train-test split & 60:40 \\
    Batch size & 32 \\ 
    Optimizer & Adam \\
    Learning rate & $10^{-3}$ \\
    Hidden dimension & 32 \\
    Epochs & 1500 \\
    \bottomrule
  \end{tabular}
  }
  \label{tab:downstream_hyperparams} 
\end{table}

\noindent \textbf{Predictive score} follows the ubiquitous ``Train on synthetic, test on real" principle by using synthetic data to train downstream models, as we require that models trained on generated data may be readily applied to real circumstances without a substantial loss of efficacy \citep{jordon2022synthetic}. The predictive score metric creates a new post-hoc predictive recurrent network, $\mathcal{P}$, which learns to predict the next observation in a sequence using normalized synthetic training data, $\{\hat{\mathbf{x}}_{1:T}^{(i)}\}_{i=1}^M$. Once the model is trained, it may then be tested using real sequences from $\{\mathbf{x}_{1:T}^{(i)}\}_{i=1}^N$, and the predictive score is evaluated based on the mean absolute error (MAE) over component prediction

\begin{align}
\text{Predictive Score} &= \frac{1}{N} \sum_{i=1}^{N} \text{MAE}\left(\mathbf{x}_{1:T}^{(i)}, \mathcal{P}\right), \label{eqn: predictive score} \\
\intertext{where,}
\text{MAE}(\mathbf{x}_{1:T}, \mathcal{P}) &= \frac{1}{nT} \sum_{t=1}^{T} \left\| \mathbf{x}_t - \mathcal{P}(\mathbf{x}_{1:t-1}) \right\|_1.
\end{align}

Here, $\lVert \cdot \rVert_1$ is the L1 norm, and $n$ is the dimensionality of sequence observations. We may observe from \eqref{eqn: predictive score} that lower is better for predictive scores. For additional details on this metric, please see \cite{yoon_time-series_2019}.

\vspace{1ex}

\noindent The \textbf{downstream AUC-ROC} from the synthetic CKD data is calculated by considering the performance of a simple GRU classifier on predicting if a CKD patient has diabetes based on the features presented in Table \ref{tab:feature_choices}. 

After a train-test split, chosen to accommodate for the high variability in CKD expression, the synthetic data was passed through one GRU layer and one fully-connected layer, followed by a sigmoid activation. The hyperparameters for the downstream model are stated in Table \ref{tab:downstream_hyperparams}.

\section{Clinician Evaluation Details} 
\label{appendix_clinician}

Figure \appendixref{fig:sample_clinician_eval} shows a typical patient profile that was shared with the clinician along with the three questions that were used to evaluate the data. 

\begin{figure}[h]
    \centering
\includegraphics[width=0.5\textwidth]{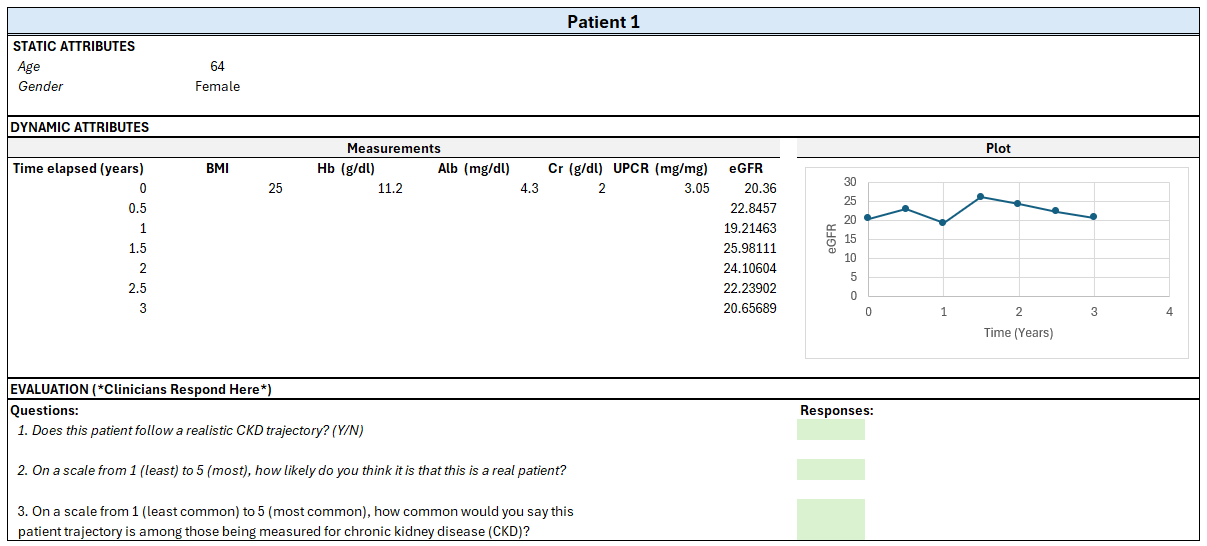}
    \caption{Sample test patient from blinded clinician evaluation.}
    \label{fig:sample_clinician_eval}
\end{figure}

We then provided the profiles to five CKD specialists, each of whom responded to the following three questions for each patient:

\begin{enumerate}
\label{list:clinician_evaluation_questions}
    \item[Q1] Does this patient follow a realistic CKD trajectory? (Y/N)
    \item[Q2] On a scale from 1 (least) to 5 (most), how likely do you think it is that this is a real patient?
    \item[Q3] On a scale from 1 (least common) to 5 (most common), how common would you say this patient trajectory is among those being measured for chronic kidney disease (CKD)?
\end{enumerate}

This judged the realism and frequency of each synthetic sample in the clinic, providing a basis for how relevant the CKD sample would be. In addition to filling out the answers to the questions within the form, clinicians provided additional written feedback.

\section{Generative Model Visualizations}
\label{appendix_viz}

In this appendix, we provide visualizations of data generated using DP-TimeGAN. To begin with, we exhibit a common test for generative time series models, which is to synthesize sinusoidal data, as it exhibits seasonality patterns that are often challenging to handle in traditional auto-regressive models. Here, we randomly sample phase and frequency values from a uniform distribution, which we use to construct a dataset of real sine waves, as described in Section \ref{datasets}. Figure~\ref{suppfig:sines_dp_timegan}
shows an example of the synthetic results from training DP-TimeGAN on the sinusoidal dataset.

As a second example, we utilize patient data from the eICU dataset. In Figures~\ref{suppfig:eicu_viz_1}, \ref{suppfig:eicu_viz_2}, and \ref{suppfig:eicu_viz_3}, we exhibit the shape of synthetic data generated using our novel Augmented-TimeGAN and DP-TimeGAN as well as the baseline models. From these graphs it is clear that visually, our novel model achieves comparable performance to TransFusion and DP Normalizing Flows for the eICU dataset. Augmented TimeGAN also performs substantially better than other benchmarks, while the DP version is not far behind.

\section{Privacy-Utility and Fidelity Tradeoff}
\label{appendix_privacy_utility}

To further describe the privacy-utility tradeoff seen with DP-TimeGAN, as mentioned in the \hyperref[limitations]{Limitations}, we measured utility and fidelity metrics with changing $\epsilon$ values. The results of this are shown in Table \ref{tab:privacy-fidelity-utility}. As shown, with increasing $\epsilon$ values, the AUC-ROC increases, and fidelity measures also become closer to optimal. Therefore, if DP-TimeGAN is used in a clinical setting, the tuning of $\epsilon$ values is necessary in order to maximize privacy while still retaining high downstream potential and accurate chronic disease deterioration.

\begin{figure*}[p]
    \centering
    
    \subfigure[Original Data]{
        \centering
        \includegraphics[width=0.48\textwidth]{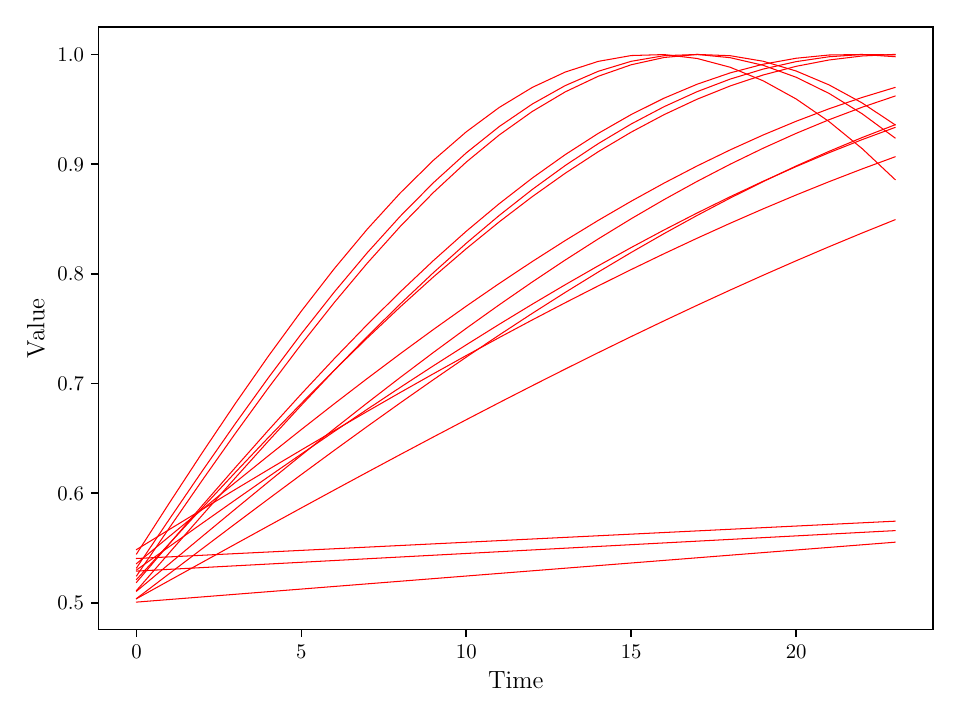}
        \label{fig:realsine_paths_dp_timegan}
        }
    \subfigure[Synthetic Data]{
        \centering
        \includegraphics[width=0.48\textwidth]{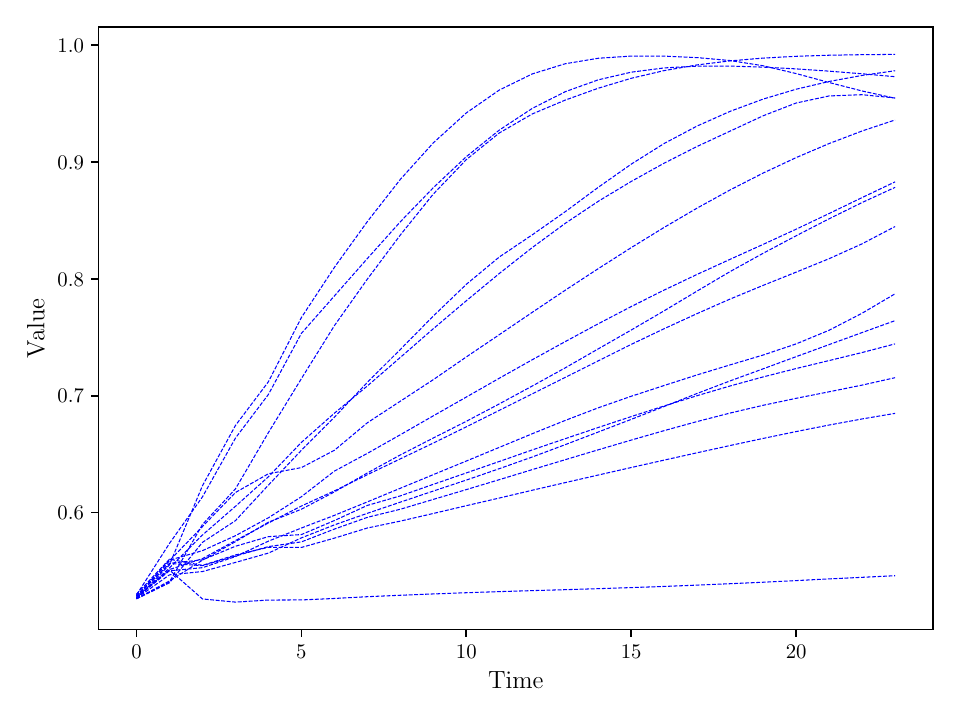}
        \label{fig:fakesine_paths_dp_timegan}
        }
    
    \vspace{0.5cm}
    
    \subfigure[PCA visualization]{
        \centering
        \includegraphics[width=0.48\textwidth]{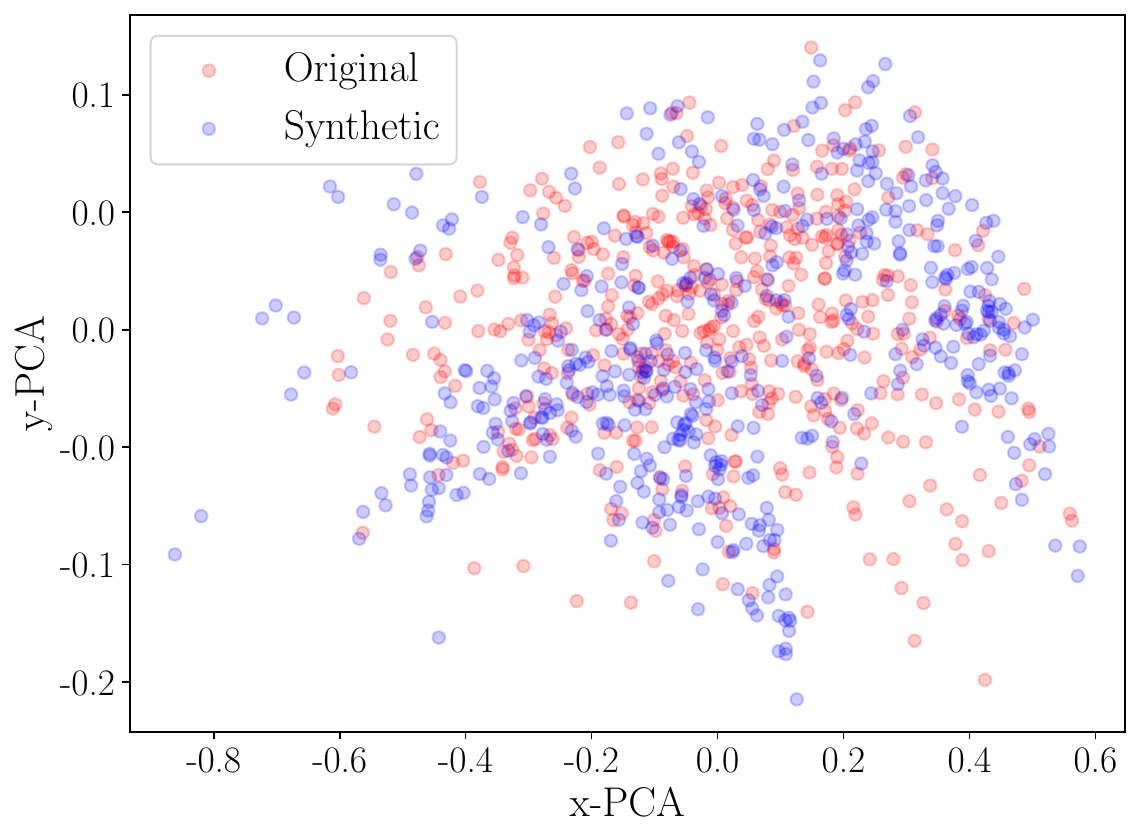}
        \label{fig:pca_sines_dp_timegan}
        }
    \subfigure[t-SNE visualization]{
        \centering
        \includegraphics[width=0.48\textwidth]{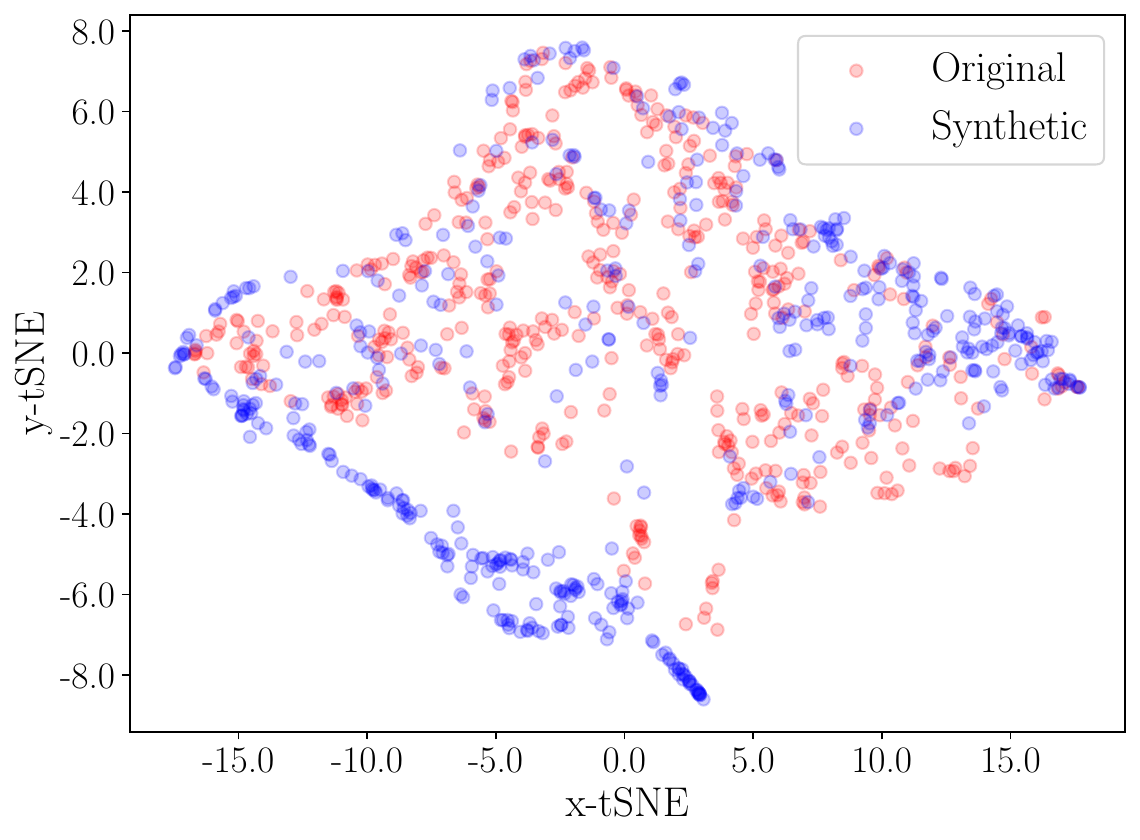}
        \label{fig:tsne_sines_dp_timegan}
        }
    \caption[Comparison of real and synthetic sinusoidal data]{Comparison of real and synthetic sinusoidal data from  DP-TimeGAN. Real data uses ($N$, $T$, $C$) = (500, 24, 5), where each feature is a randomly generated sine wave; training parameters are: $\# \text{epochs}= 7000$, $\# \text{layers} = 3$, $\text{latent-dim} = 24$, $\gamma=1$. Plots (a) and (b) isolate one feature of real and synthetic sine waves, respectively; plots (c) and (d) compare the PCA and t-SNE results, respectively, for the two datasets.}
    \label{suppfig:sines_dp_timegan}
\end{figure*}

\begin{figure*}[p]
    \begin{tabular}{@{}cc@{}}
        
        \includegraphics[width=0.45\textwidth]{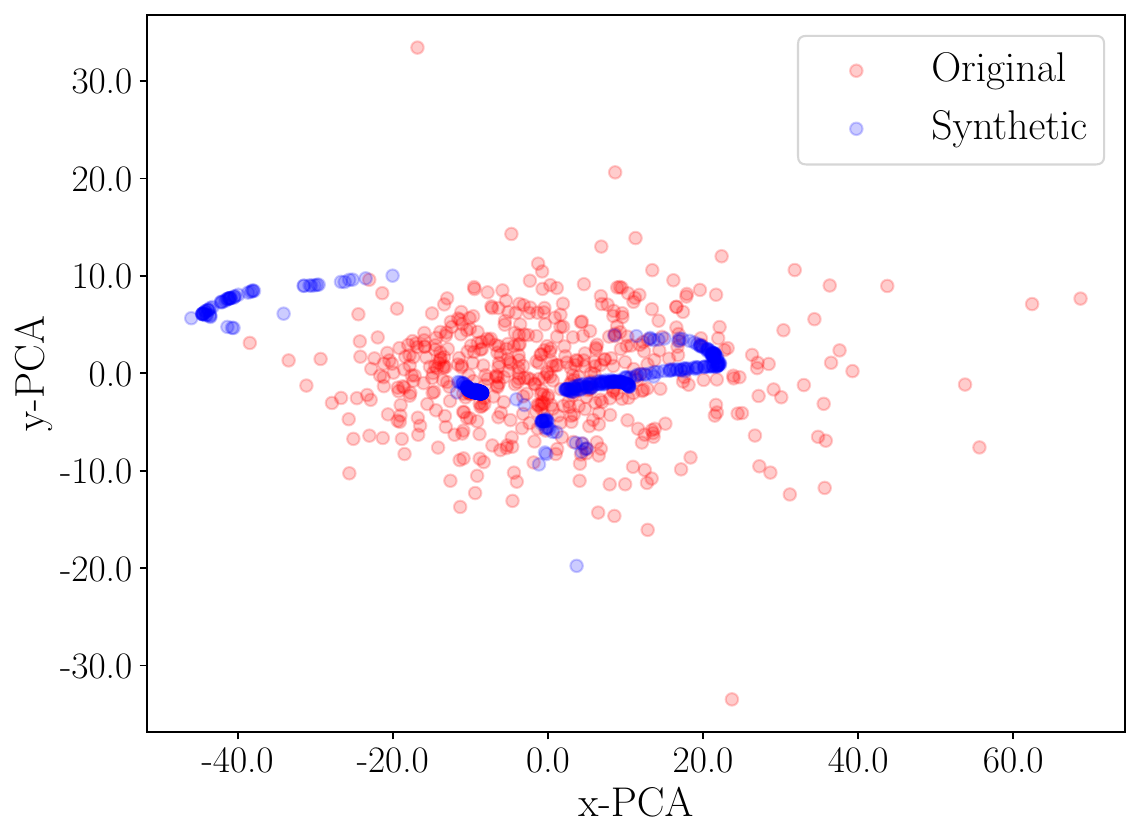}
        \label{fig:pca_eicu_reg_timegan}
        &
        \includegraphics[width=0.45\textwidth]{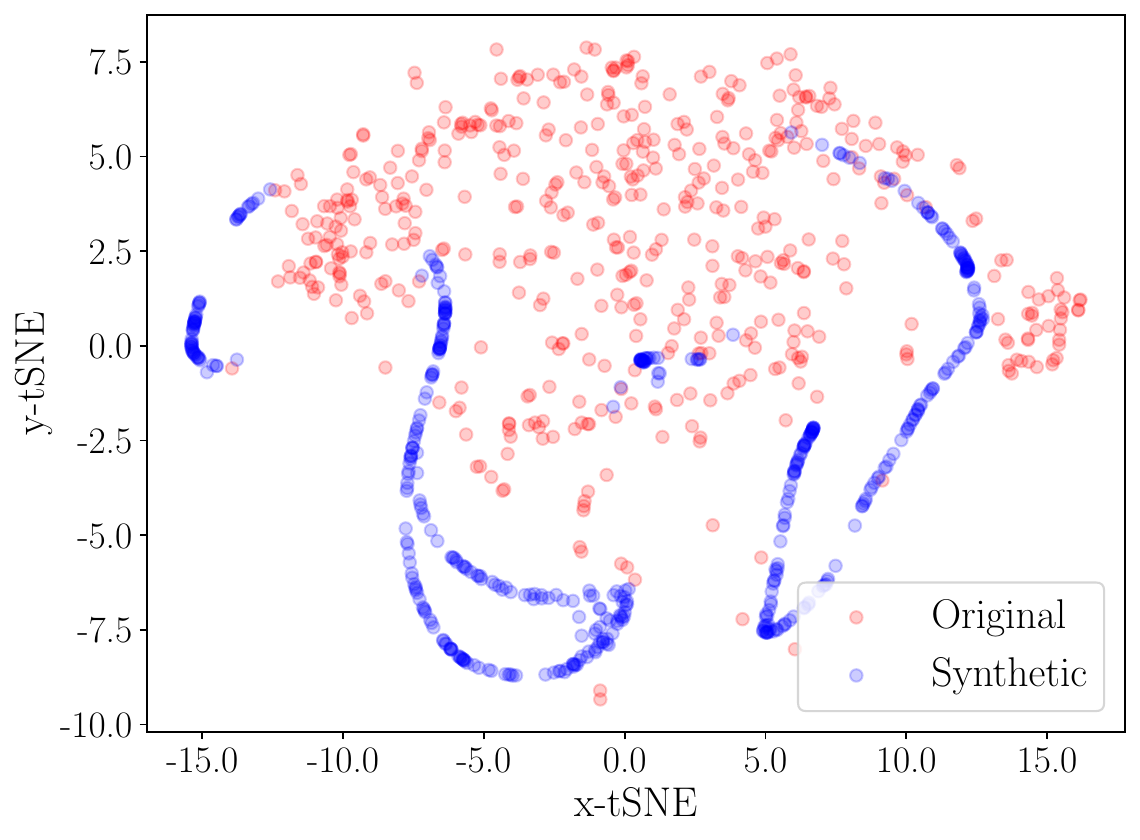}
        \label{fig:tsne_eicu_reg_timegan}
        \\

        \includegraphics[width=0.45\textwidth]{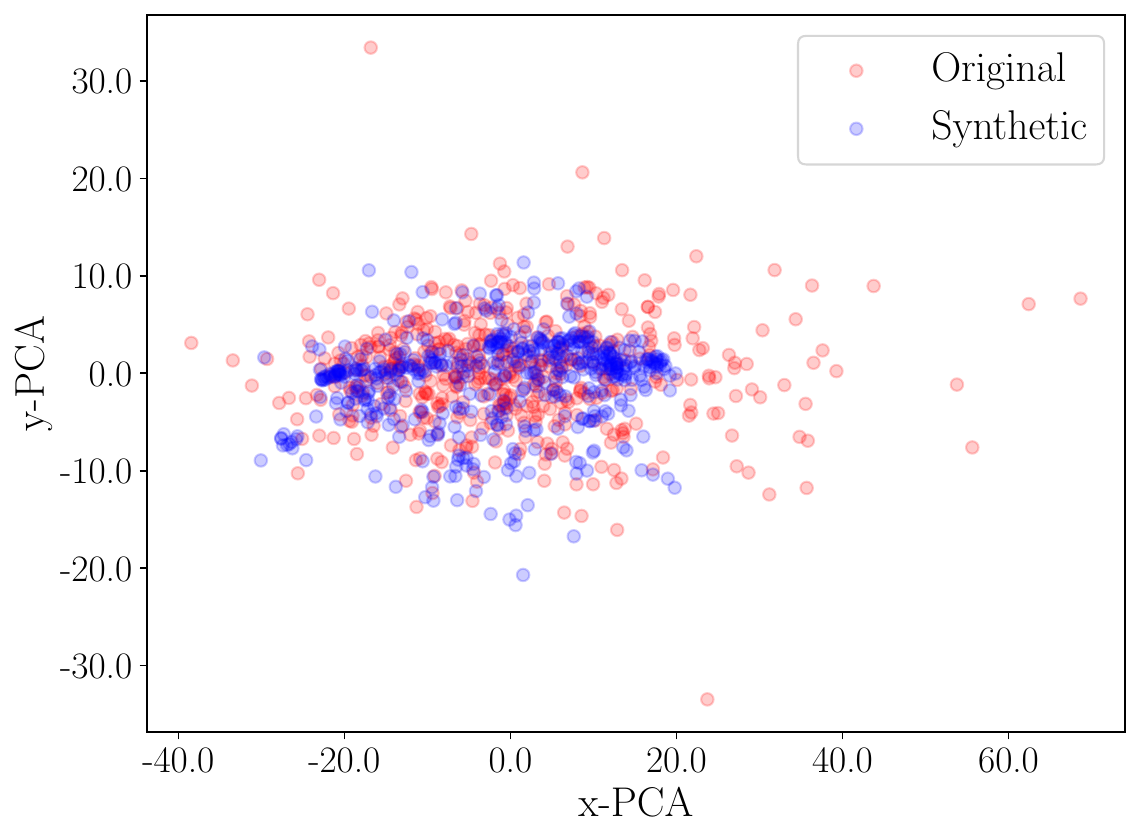}
        \label{fig:pca_eicu_aug_timegan}
        &
        \includegraphics[width=0.45\textwidth]{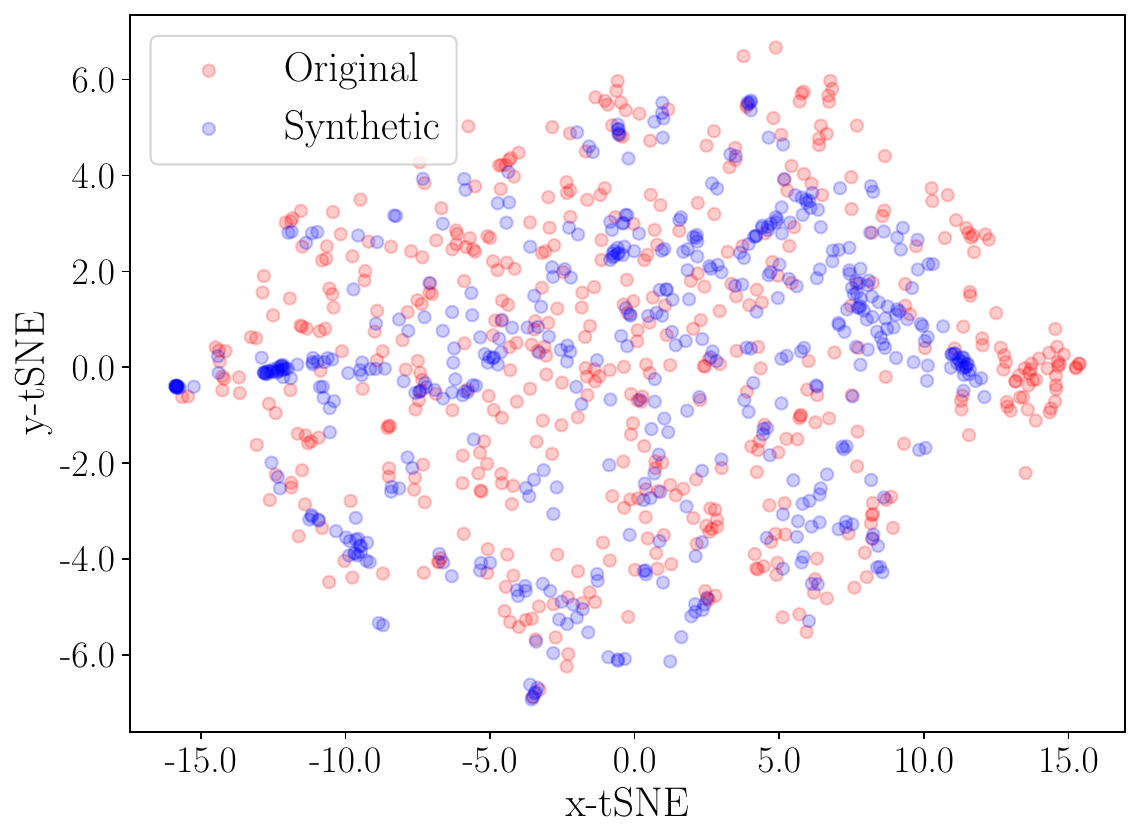}
        \label{fig:tsne_eicu_aug_timegan}
        \\

        \includegraphics[width=0.45\textwidth]{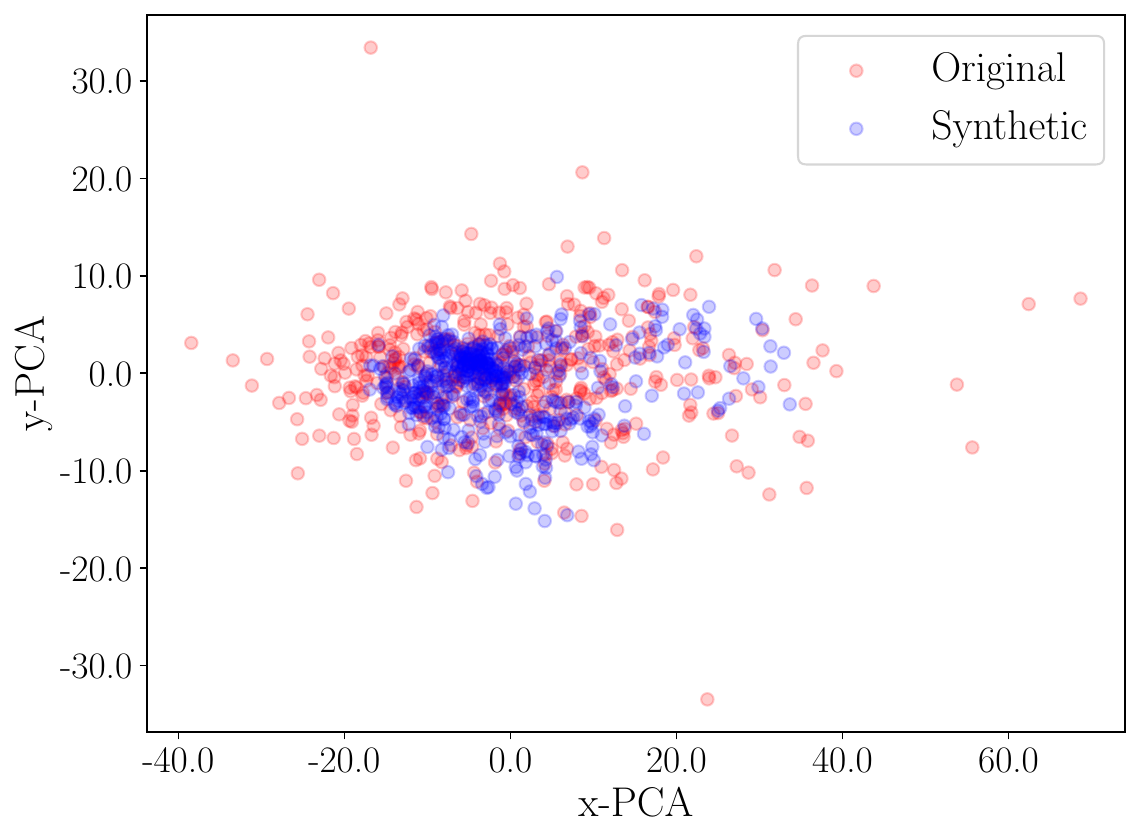}
        \label{fig:pca_eicu_dp_timegan}
        &
        \includegraphics[width=0.45\textwidth]{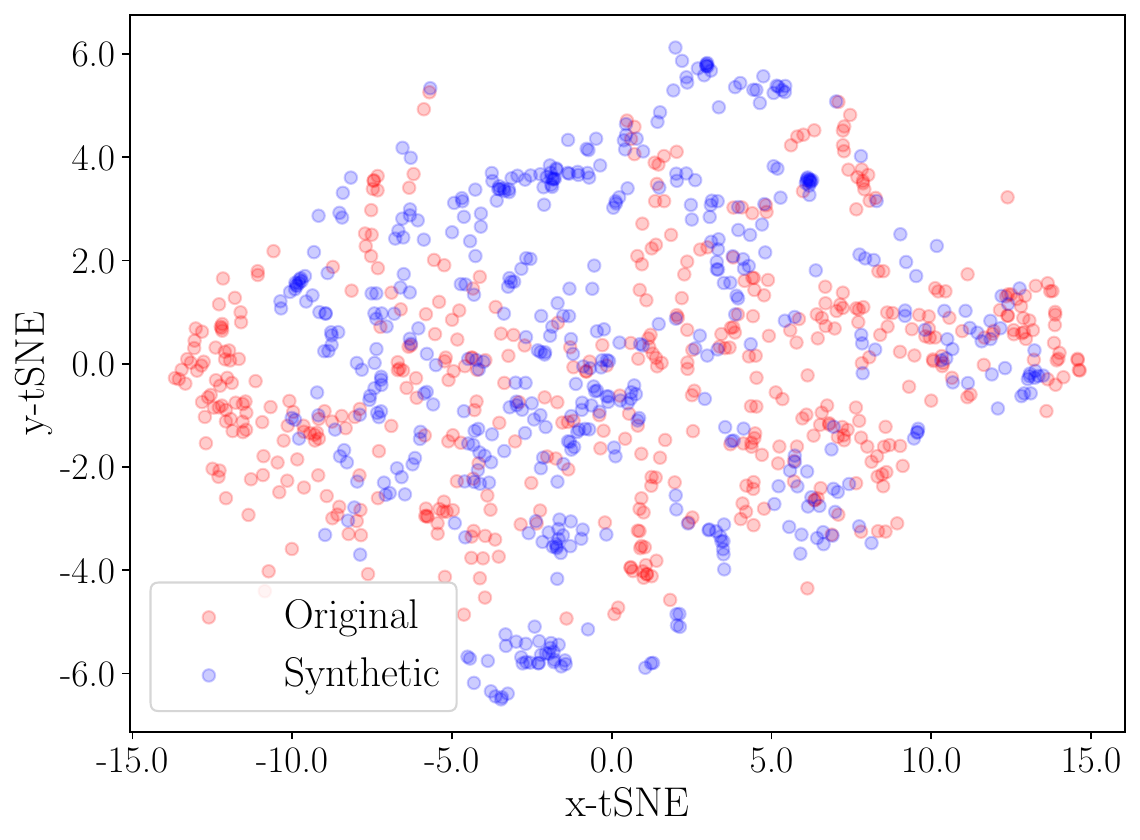}
        \label{fig:tsne_eicu_dp_timegan}
        \\

    \end{tabular}   
    \caption[Comparison of real and synthetic eICU data]{Comparison of real and synthetic eICU data from the Regular TimeGAN (Row 1), Augmented TimeGAN (Row 2), and DP-TimeGAN (Row 3) models. TimeGAN models utilize the same parameters as mentioned in Table \ref{tab:all_datasets_metrics}. Plots in the first and second columns from the left compare the PCA and t-SNE results, respectively.}
    \label{suppfig:eicu_viz_1}
\end{figure*}

\begin{figure*}[p]
    \begin{tabular}{@{}cc@{}}

        \includegraphics[width=0.45\textwidth]{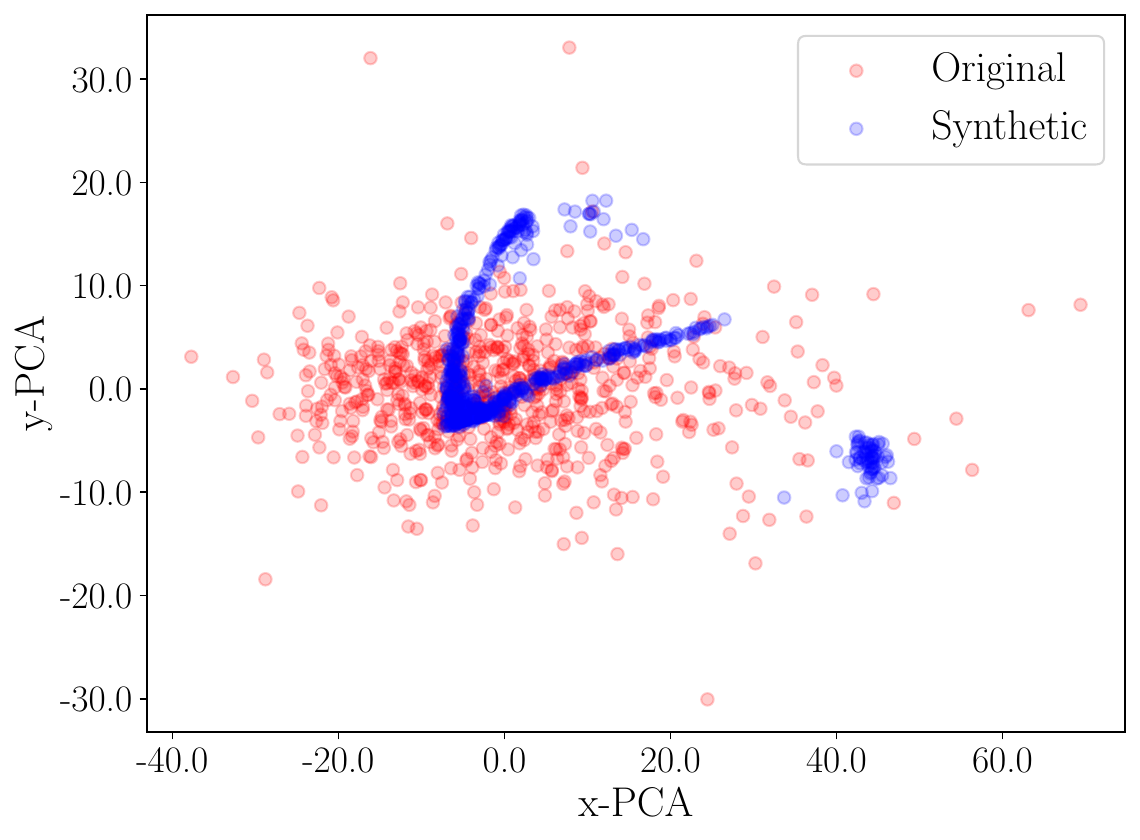}
        \label{fig:pca_eicu_seriesgan}
        &
        \includegraphics[width=0.45\textwidth]{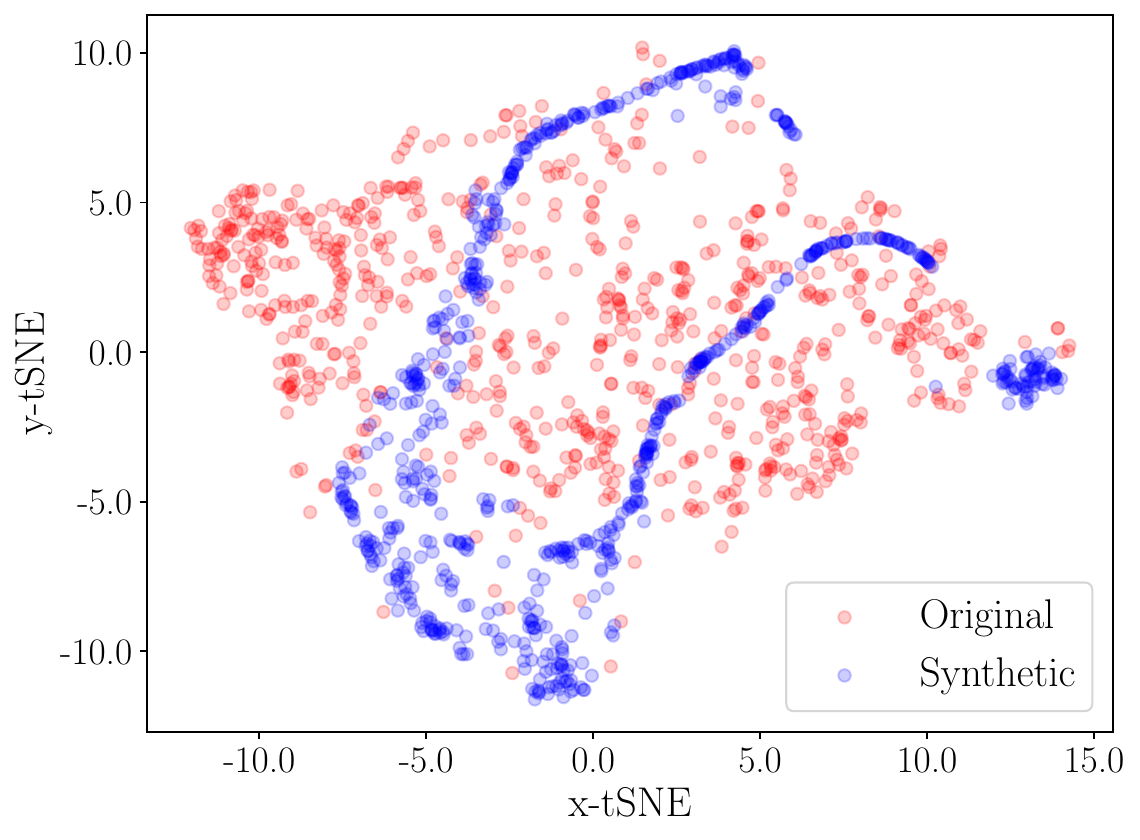}
        \label{fig:tsne_eicu_seriesgan}
        \\
        
        \includegraphics[width=0.45\textwidth]{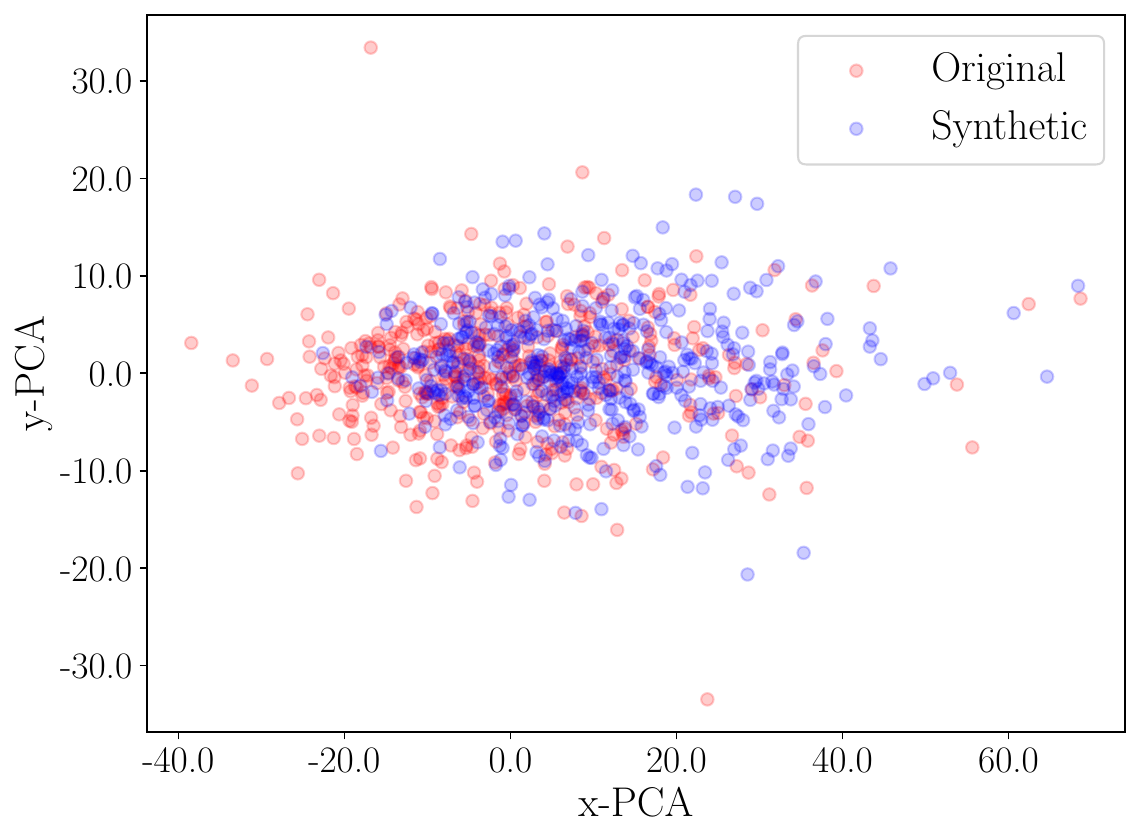}
        \label{fig:pca_eicu_transfusion}
        &
        \includegraphics[width=0.45\textwidth]{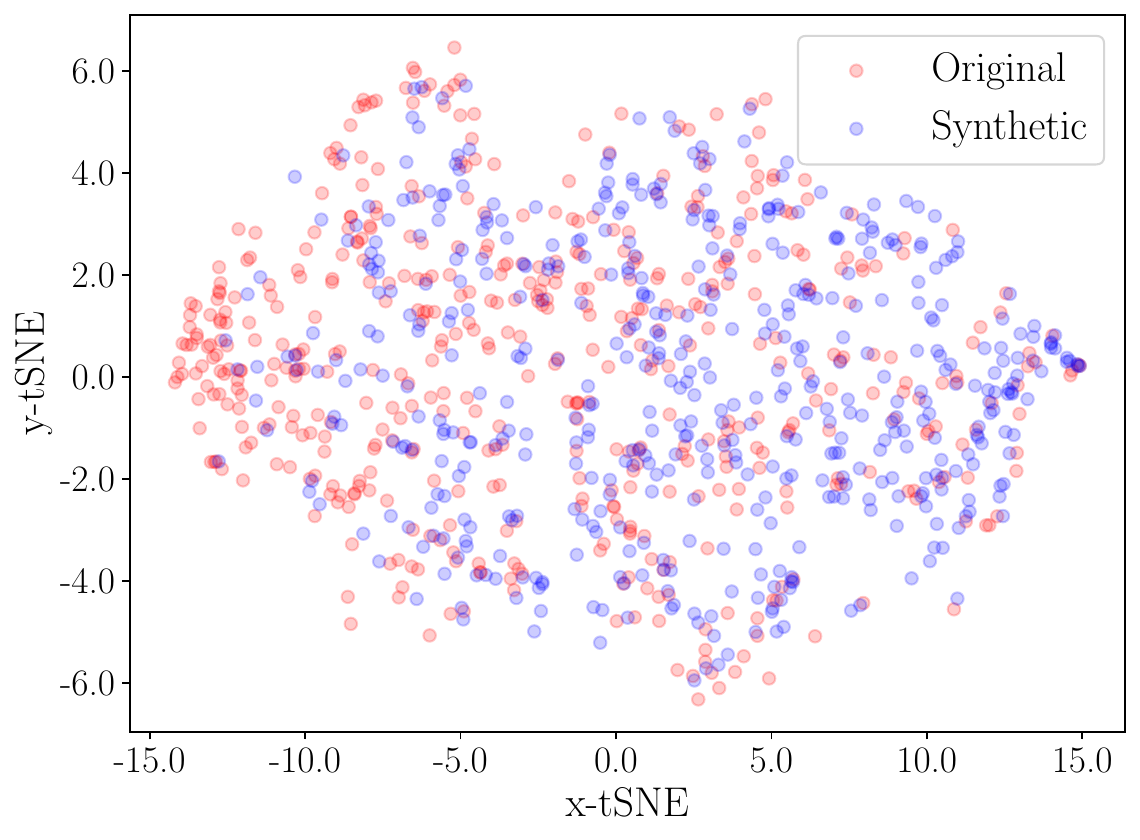}
        \label{fig:tsne_eicu_transfusion}

        \\

        \includegraphics[width=0.45\textwidth]{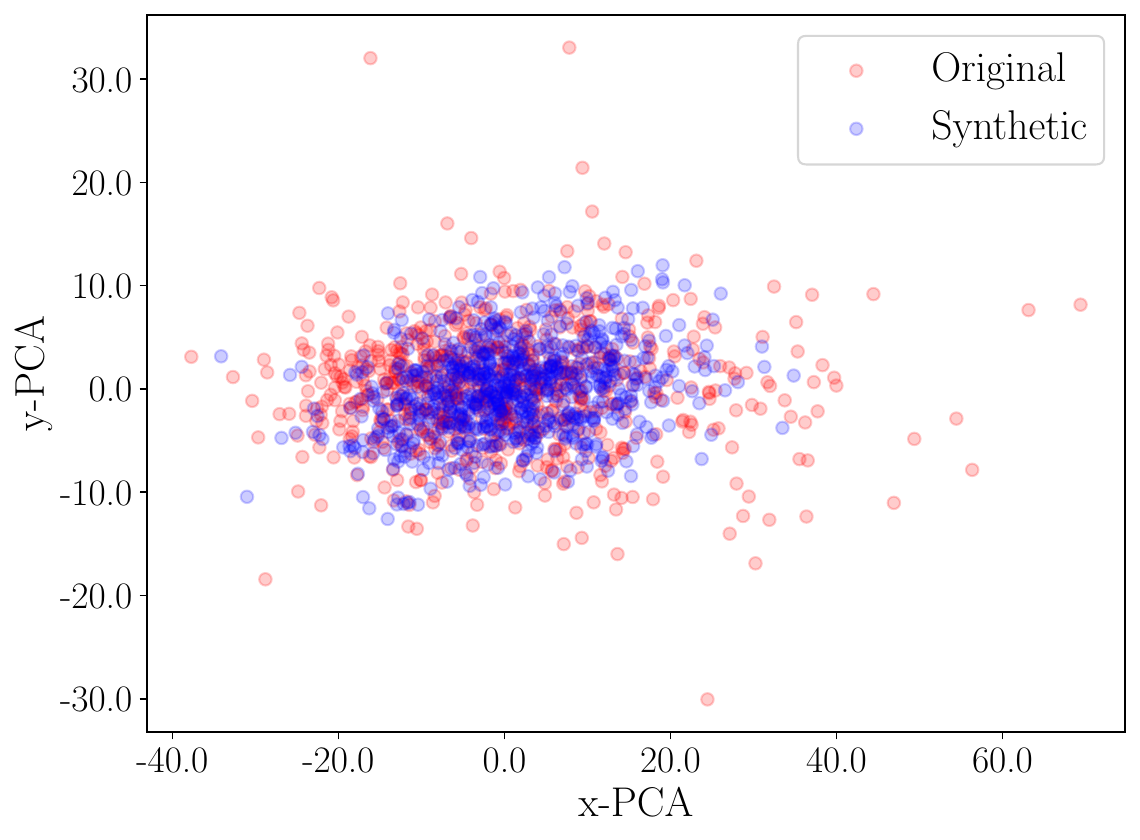}
        \label{fig:pca_eicu_dpnf}
        &
        \includegraphics[width=0.45\textwidth]{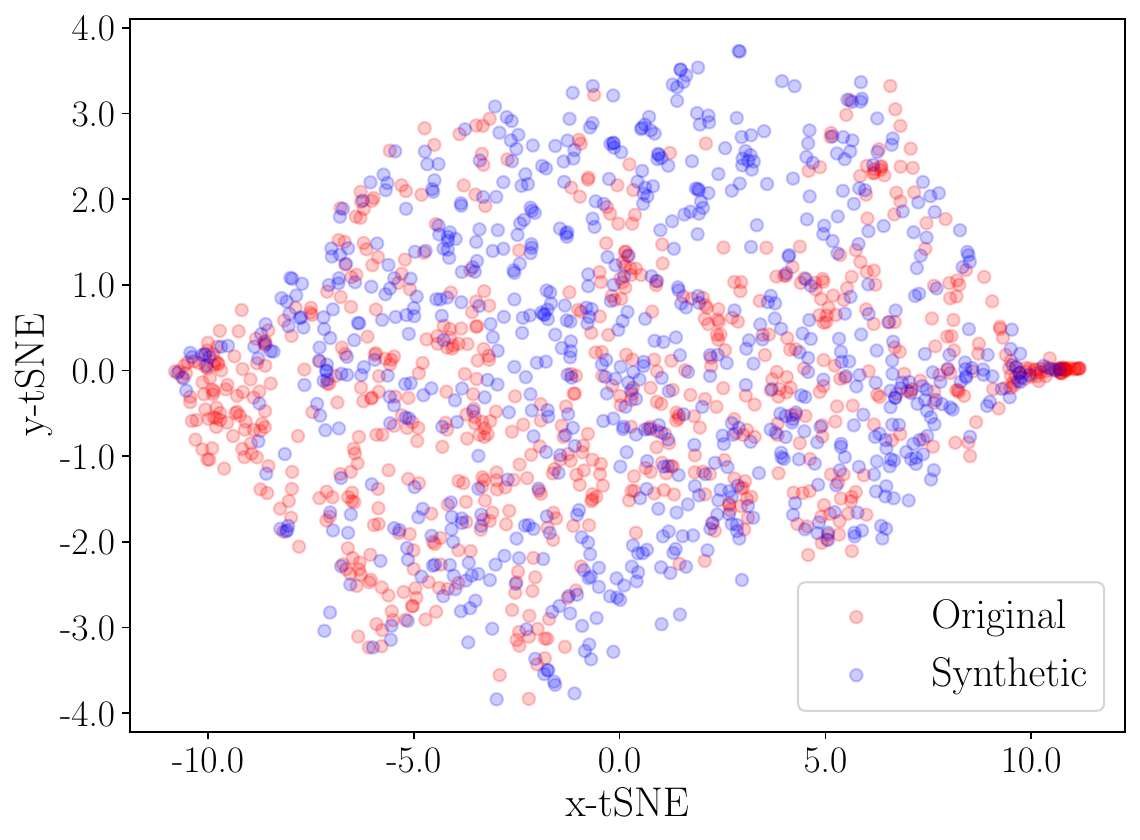}
        \label{fig:tsne_eicu_dpnf}
        
        \\       
    \end{tabular}   
    \caption[Comparison of real and synthetic eICU data]{Continued comparison of real and synthetic eICU data from the SeriesGAN (Row 1), TransFusion (Row 2), and DP Normalizing Flows (Row 3) models. Benchmark models replicate the hyperparameters from their respective publications. Plots in the first and second columns from the left compare the PCA and t-SNE results, respectively.}
    \label{suppfig:eicu_viz_2}

\end{figure*}

\begin{figure*}[t]
    \begin{tabular}{@{}cc@{}}

        \includegraphics[width=0.45\textwidth]{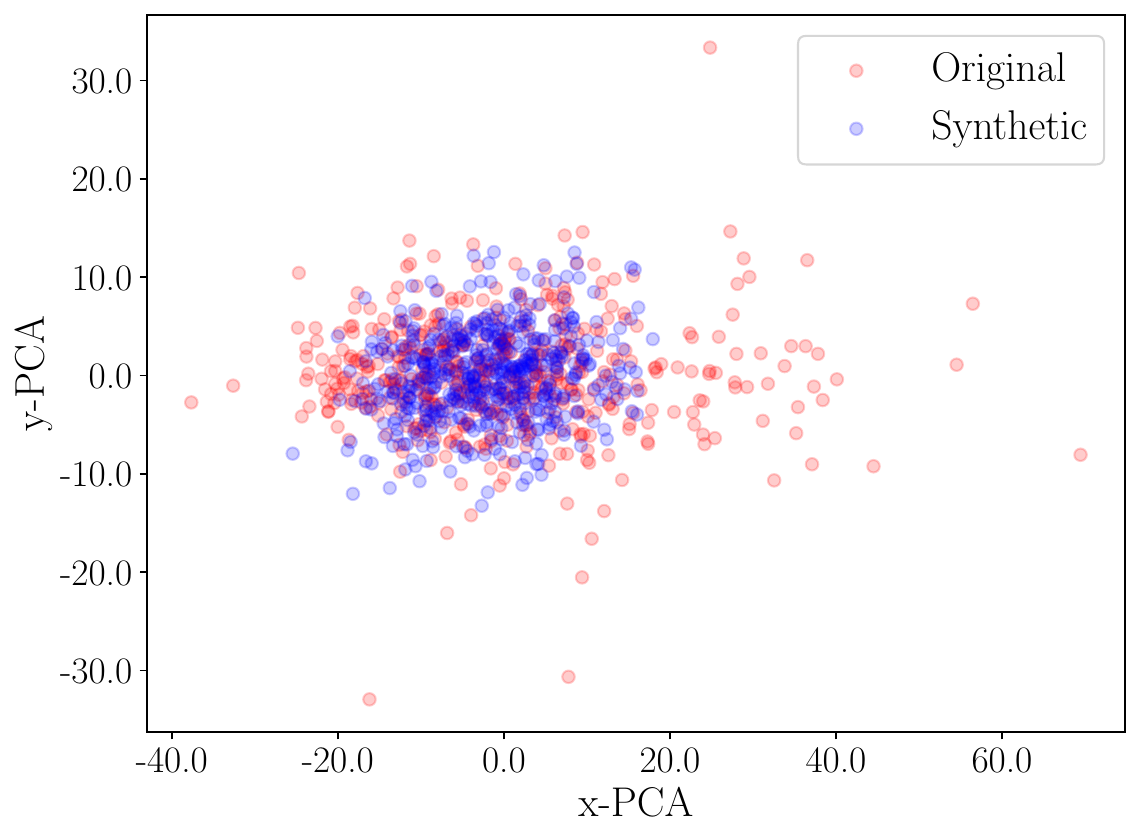}
        \label{fig:pca_eicu_timediff}
        &
        \includegraphics[width=0.45\textwidth]{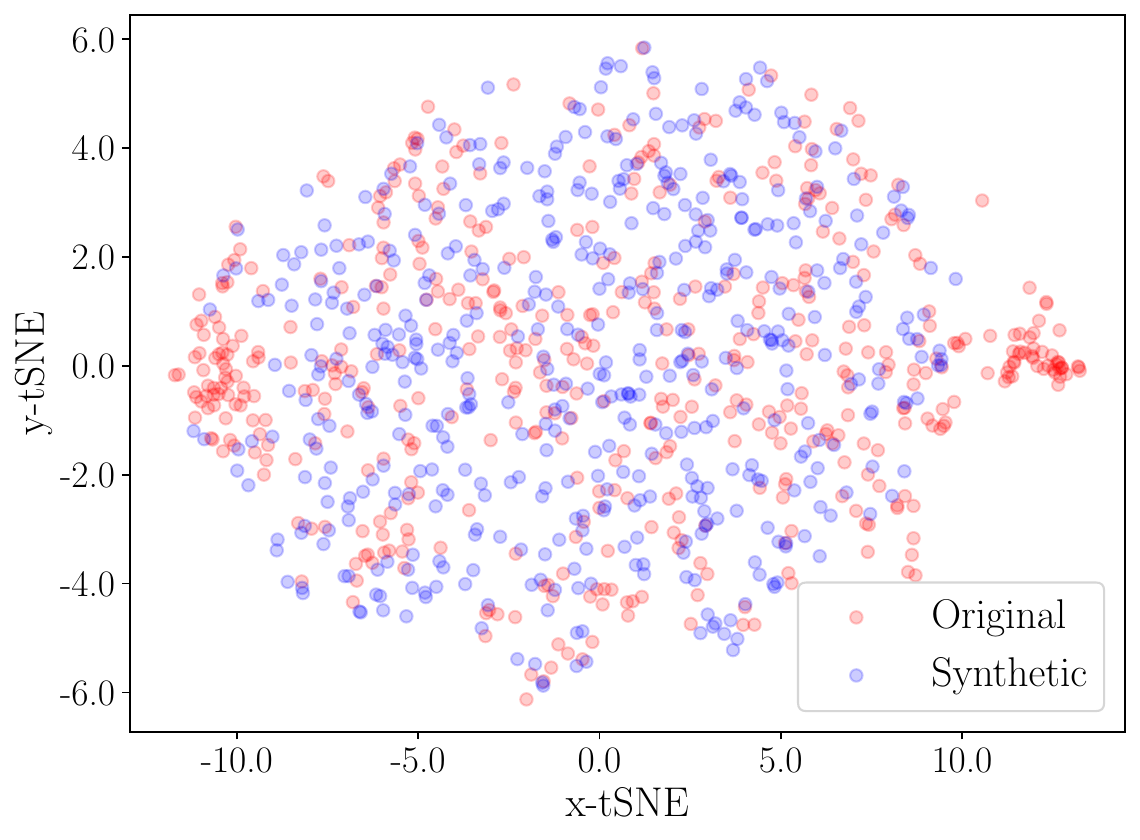}
        \label{fig:tsne_eicu_timediff}
        
        \\       
    \end{tabular}   
    \caption[Comparison of real and synthetic eICU data]{Continued comparison of real and synthetic eICU data from the TimeDiff model. Benchmark models replicate the hyperparameters from their respective publications. Plots in the first and second columns from the left compare the PCA and t-SNE results, respectively.}
    \label{suppfig:eicu_viz_3}
\end{figure*}

\begin{table*}[h]
  \centering 
  \caption{Privacy-fidelity and -utility study of DP-TimeGAN on the CKD dataset. DP-TimeGAN utilizes the same parameters as mentioned in Table \ref{tab:all_datasets_metrics}.} 
  \scalebox{0.9}{
  \begin{tabular}{>{\raggedright\arraybackslash}p{1.2cm} > {\raggedright\arraybackslash}p{2cm} >{\raggedright\arraybackslash}p{2cm} > {\raggedright\arraybackslash}p{2cm} > {\raggedright\arraybackslash}p{2cm} }
  \toprule
    \textbf{Epsilon} & \textbf{MMD} & \textbf{DS} & \textbf{$\alpha$-precision} & \textbf{Downstream AUC-ROC}\\
    \midrule
    10 & $0.122 \pm 0.058$ & $0.380 \pm 0.040$ & $0.852 \pm 0.051$ & $0.501 \pm 0.025$ \\
    20 & $0.091 \pm 0.046$ & $0.312 \pm 0.053$ & $0.844 \pm 0.035$ & $0.549 \pm 0.063$ \\ 
    30 & $0.069 \pm 0.002$ & $0.320 \pm 0.076$ & $0.840 \pm 0.079$ & $0.553 \pm 0.074$ \\
    40 & $0.079 \pm 0.050$ & $0.321 \pm 0.057$ & $0.870 \pm 0.032$ & $0.553 \pm 0.019$ \\
    50 & $0.064 \pm 0.002$ & $0.297 \pm 0.030$ & $0.894 \pm 0.066$ & $0.577 \pm 0.026$ \\
    \bottomrule
  \end{tabular}
  }
  \label{tab:privacy-fidelity-utility} 
\end{table*}

\end{document}